\PassOptionsToPackage{table,dvipsnames}{xcolor}
\documentclass{muxlab}
\usepackage{amsmath,amssymb,mathtools,nicefrac}
\usepackage{algorithm}
\usepackage{algpseudocode}
\usepackage{amsmath,amssymb}
\usepackage{wrapfig}

\newtheorem{theorem}{Theorem}
\newtheorem{proposition}{Proposition}
\newtheorem{lemma}{Lemma}
\newtheorem{remark}{Remark}

\definecolor{softgreen}{RGB}{239,247,239}

\newcommand{\questionbox}[1]{\par\smallskip\noindent\begingroup
\setlength{\fboxsep}{6pt}\colorbox{softgreen}{\begin{minipage}{\dimexpr\linewidth-2\fboxsep\relax}
\textbf{Question.}\enspace #1\end{minipage}}\endgroup\par\smallskip}

\newenvironment{proof}[1][Proof]{\par\noindent\textit{#1.} }{\hfill$\square$\par}

\renewcommand*{\backref}[1]{}
\renewcommand*{\backrefalt}[4]{\ifcase#1\relax\or(page #2)\else(pages #2)\fi}
\definecolor{BrandGreen}{HTML}{00754B}
\definecolor{BrandRed}{HTML}{B93430}
\definecolor{muxviolet}{HTML}{00754B}
\definecolor{muxlight}{HTML}{D7EBE2}
\definecolor{muxaccent}{HTML}{00754B}
\definecolor{muxdark}{HTML}{174A35}
\definecolor{muxrule}{HTML}{B4D4C3}
\definecolor{muxkeybox}{HTML}{F2F8F4}
\definecolor{muxlink}{HTML}{00754B}
\hypersetup{citecolor=BrandGreen,linkcolor=BrandGreen,urlcolor=BrandGreen}
\title{Circular Predictive Coding Unlocks Effective Depth}
\renewcommand{\authorlist}{%
 {\sffamily\color{muxfg}\textbf{Tommaso Salvatori}\textsuperscript{1,2},\;\;\textbf{Matteo Forasassi}\textsuperscript{1,2}}%\\[2pt]%
 }
\renewcommand{\affiliationlist}{%
 {\small\textsuperscript{1}Relational Intelligence, Inc\quad\textsuperscript{2}Vienna University of Technology}\\[2pt]%
 {\small\texttt{\{tommaso,matteo\}@rel-int.ai}}}
\abstract{
Predictive coding networks perform credit assignment through an energy minimization process. Despite their ability to train models with arbitrary network topologies, they have mostly been used to train feedforward architectures to match the performance of backpropagation. However, they fail to do so when the models become too deep, because error signals become unstable or vanish as they travel through many layers. We address this limitation with circular networks, which get their depth directly from the energy minimization, rather than from stacking layers. Formally, we prove that convergence to an equilibrium of the energy functional of a circular predictive coding network approximates the fixed-point iteration process of a deep equilibrium model, showing that circularity alone suffices to unlock effective depth. Empirically, we show that circular models (i) outperform other predictive coding models of the same size on image classification tasks, matching the accuracy of state-of-the-art predictive coding ResNets and VGG networks despite having substantially fewer parameters; (ii) solve constraint satisfaction problems that require many steps of iterative reasoning, a task not feasible with feedforward models at all, reaching $99.3\%$ test accuracy on Sudoku-hard via nested energy minimization loops; and (iii) scale to autoregressive language modeling by training a circular transformer on TinyStories and WikiText-103, before out of reach for predictive coding models.  These results have direct implications for physical neural networks, where computational depth can be traded for recursion on a fixed-size circuit.

\par\medskip
{\centering
  \small\sffamily
  \textcolor{BrandGreen}{%
    \textbf{Code:}\enspace
    \href{https://github.com/rel-int/CircularEP}{%
      \textcolor{BrandGreen}{\nolinkurl{https://github.com/rel-int/CircularEP}}}%
    \quad
    \ \ \ \ \ 
    \textbf{Demo:}\enspace
    \href{https://rel-int.github.io/CircularEP/}{%
      \textcolor{BrandGreen}{\nolinkurl{https://rel-int.github.io/CircularEP/}}}%
  }%
  \par

}

}

\begin{document}
\maketitle

% \begin{figure}[H]
% \centering
% \includegraphics[width=\linewidth]{CircularEP.pdf}
% %\vspace{-0.5cm}
% \caption{(a) An example of a PC graph. Zoomed in, the dynamics of every vertex, divided into value nodes, predictions, and error nodes; (b) An example of a feedforward predictive coding model; (c) The same feedforward model, where we have identified the first and last value nodes to create a circular structure, that will allow the model to learn and solve reasoning tasks.
% }
% \label{fig:models_intro}
% \end{figure}

\begin{figure}[H]
\centering
\includegraphics[width=0.97\linewidth]{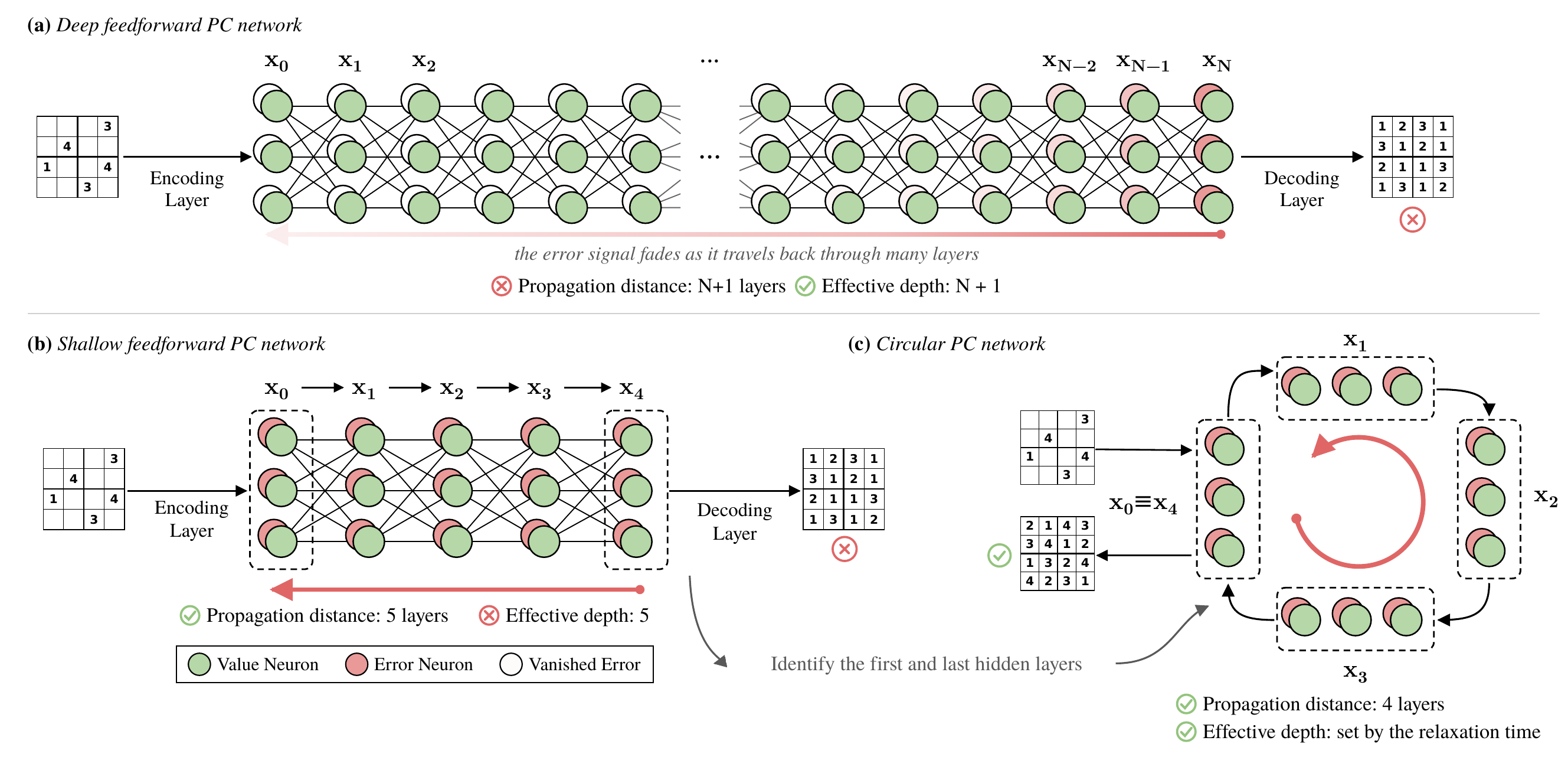}
%\vspace{-0.5cm}
\caption{Every vertex of a predictive coding network has a value neuron, and an error neuron. (a)~In a deep feedforward network, the error created at the output must travel back through $N$ layers. It fades along the way, so the early layers get almost no learning signal. (b)~A shallow network propagates the error well, but it is not deep enough to solve the task. (c)~Identifying the first and last hidden layers ($\mathbf{x}_0 \equiv \mathbf{x}_4$) closes the shallow network into a loop. The error now circulates through at most four layers, while the effective depth is set by how long the energy relaxes, so the network can solve tasks that need many steps of reasoning.}
\label{fig:models_intro}
\end{figure}

% \begin{figure}[H]
% \centering
% \includegraphics[width=\linewidth]{CircularEP.pdf}
% %\vspace{-0.5cm}
% \caption{(a) An example of a PC graph. Zoomed in, the dynamics of every vertex, divided into value nodes, predictions, and error nodes; (b) An example of a feedforward predictive coding model, divided into value nodes and error nodes; (c) The same feedforward model, where we have identified the first and last value nodes to create a circular structure, that will allow the model to learn and solve reasoning tasks.
% }
% \label{fig:models_intro}
% \end{figure}

\section{Introduction}

Predictive coding (PC) \citep{rao1999predictive,friston2005theory} is a brain-inspired learning algorithm that performs credit assignment by relaxing an energy functional defined over neurons. Its computations are local in space and time, so it can in principle train architectures of any topology, including looped ones \citep{salvatori2022learning}. Yet, these properties have rarely been exploited in practice, as recent work has largely used this relaxation to reach the equilibrium needed for credit assignment of feedforward architectures with a fixed number of layers \citep{pinchetti2025benchmarking}. This choice was pragmatic: Since predictive coding approximates backprop on feedforward networks, the architectures, initializations and optimization techniques developed in deep learning could be reused almost unchanged, which allowed a small community to scale PC from MNIST to ImageNet in a few years \citep{whittington2017approximation,kerjan2026training}. The downside, however, is that research directions such as the study of novel architectures and different ways to exploit the energy minimization process have been overlooked, which in the long term could help solve some of the underlying problems of this algorithm.

The main problem of predictive coding models is depth: their performance degrades once they grow beyond roughly seven layers, because the energy of the model concentrates near the output and early layers receive almost no learning signal \citep{pinchetti2025benchmarking,innocenti2026mu}. Recent work has heavily focused on addressing this by modifying the learning algorithm, with mixed results: They have either formally tackled the problem, but have only shown results on MNIST-scale tasks \citep{innocenti2026mu,seely2026augmented}, or have reached strong performance on CIFAR10/100, by modifying the algorithm to mitigate the symptom of this energy imbalance, rather than remove the cause \citep{ha2026stable,qi2025towards}. Here, we address this architecturally by using the core properties of predictive coding: locality of computations, that allows the training of models with looped topologies, and message passing schemes that run until convergence to a stationary point of the energy function. Specifically, we focus on the following question:

\questionbox{Can energy-based relaxation be turned into a form of effective depth, and what architectural structure is required for such behavior to emerge?}

Our answer is to close a shallow feedforward PC network into a loop, by identifying its first and last hidden layers. In the resulting \emph{circular predictive coding network}  (CPCN), effective depth is set by how long the energy relaxes, not by how many layers are stacked, while the error signal only travels around a loop of at most four hidden layers. Closing the loop, however, breaks the two correspondences that allowed PC to reuse the machinery of backprop.
%in inference, the free phase converges to the activations of a forward pass; in learning, the weight updates approximate the gradients of backprop \citep{whittington2017approximation,millidge2020predictive}. Both of these useful correspondences have been vital towards the scaling up of predictive coding networks to complex tasks. 
Luckily, there is another literature from which we can take inspiration: that of deep equilibrium models (DEQs) \citep{bai19}, whose output is the fixed point of a single transformation, and whose parameters are trained via the implicit function theorem (IFT). We will show how both correspondences can be re-derived in this case, using recent work that connects predictive coding to equilibrium propagation \citep{kerjan2026training}, and equilibrium propagation to the IFT \citep{zucchet2022beyond}. Our contributions are briefly as follows:
\begin{itemize}[leftmargin=*]
    \item We formally prove that in CPCNs trained via predictive coding (i) the free phase has a unique zero-energy equilibrium, which coincides with the fixed point of an equivalent DEQ; and (ii) building on \citep{zucchet2022beyond}, training a CPCN with EP approximates the implicit gradient of this DEQ. As such results are either approximations, or may not hold under machine precision, we verify them empirically in GPU simulations;
    \item We test CPCNs on image classification tasks, showing that it reaches performance comparable to much larger PC-based ResNet18 models on CIFAR10/100, using substantially fewer parameters, and scales to ImageNet32
    . We further provide evidence that the trained model uses the effective depth of the loop, as multiple test images converge to the correct prediction only after tens of different forward passes. 
    \item We test CPCNs on Sudoku-hard, the first application of PC to constraint satisfaction tasks. Here, we observe adaptive computation properties, as harder puzzles are solved by allocating more relaxation steps than the ones the model was trained on. We further introduce a novel training scheme, made of multiple sequential energy minimization processes and inspired by HRM and TRM \citep{wang2025hierarchicalreasoningmodel,jolicoeur2025less}, which reaches $99.3\%$ board accuracy against $97\%$ of the standard formulation, with a model $10\times$ smaller. 
    \item We train a recursive transformer on autoregressive language modelling on the TinyStories and WikiText-103 datasets \citep{eldan2023tinystories,merity2017wiki103}, the largest NLP datasets tested with PC to date, matching the performance of a DEQ of the same size.
\end{itemize}

%This mirrors how training is defined in This is analogous to fixing the potentials of some nodes of a resistive network with voltage sources. This mirrors measuring the potential at other nodes of the circuit, or the displacement of other nodes of the elastic network, once the system has settled. 
%
%$\dot{\mathbf{x}}_{\mathrm{free}} = -\sfrac{\partial E}{\partial\mathbf{x}_{\mathrm{free}}}$. A free equilibrium $\mathbf{x}^0$ is 

\begin{figure}[t]
\centering
\includegraphics[width=\linewidth]{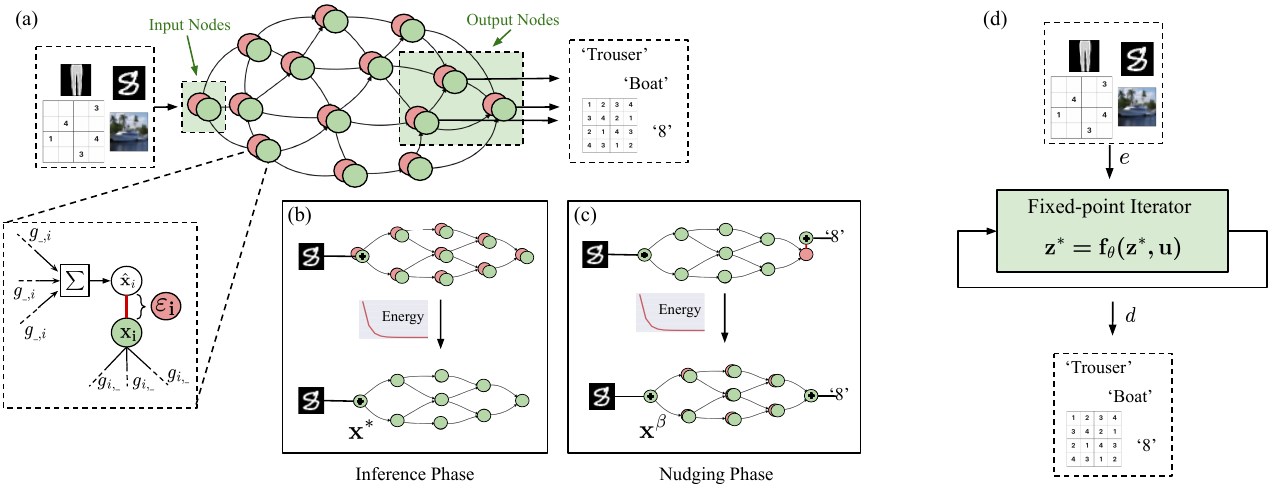}
%\vspace{-0.5cm}
\caption{(a) An example of a PC graph. Zoomed in, the dynamics of every vertex, divided in value nodes, predictions, and error nodes; (b) Inference phase on a directed acyclic graph, where the energy minimization converges to a state of zero error; (c) Corresponding nudged phase, where the value node of the output is nudged towards the label, and the error corresponding to the loss term is distributed through the network via the relaxation process; (d) a sketch of a deep equilibrium model.
}
\label{fig:models}
\end{figure}

\section{Background}

%As the goal is to prove an equivalence between deep equilibrium models \citep{bai19} and our circular formulation of predictive coding, in this section we review both concepts. We start from \emph{predictive coding graphs}, a generalization of predictive coding networks that can be used to describe models with arbitrary topologies \citep{salvatori2022learning}.

\subsection{Predictive Coding Graphs}

\emph{Predictive coding graphs} are a generalization of predictive coding networks that can be used to describe models with arbitrary topologies \citep{salvatori2022learning}. Let $\mathcal{G}=(\mathcal{V},\mathcal{E})$ be a directed graph, whose vertices we index by $i\in\mathcal{V}$, and whose edges  $(j,i)\in\mathcal{E}$ carry a parametric map $g_{j,i}$. Every vertex encodes three quantities: a random variable called value node $\mathbf{x}_i\in\mathbb{R}^{d_i}$, which represents the current neural activity of the vertex; a \emph{prediction}
\begin{equation}
    \hat{\mathbf{x}}_i = \sum_{j\in\mathrm{pa}(i)} g_{j,i}(\mathbf{x}_j),
\end{equation}
computed using the value nodes of its parents along the incoming edges; and an error node $\boldsymbol{\varepsilon}_i = \mathbf{x}_i-\hat{\mathbf{x}}_i$, defined as the mismatch between the activity of the vertex and the signals it receives. Collecting all value nodes into a single vector $\mathbf{x}=(\mathbf{x}_i)_{i\in\mathcal{V}}$, and using $\theta$ to represent the joint parameters of the parametric maps, the system is governed by an energy given by the sum of the squared prediction errors, which results in the following Lyapunov function for the dynamics:
\begin{equation}
    E(\theta,\mathbf{x})
    = \frac{1}{2}\sum_{\mathcal{V}}
      \|\mathbf{x}_i - \hat{\mathbf{x}}_i\|_2^2
    = \frac{1}{2}\sum_{\mathcal{V}}
      \|\boldsymbol{\varepsilon}_i\|_2^2.
    \label{eq:energy}
\end{equation}
\paragraph{Inference.}
As PC graphs can be seen as a physical system governed by the above energy function, it is possible to present an input signal $\mathbf{u}$ to the model by imposing boundary conditions on a subset of input vertices $\mathcal{I}\subset\mathcal{V}$, that is, by clamping $\mathbf{x}_{\mathcal{I}}=\mathbf{u}$. The remaining value nodes are then updated following  the gradient flow of the energy until they reach a free equilibrium $\mathbf{x}^*$, that is, a state at which $\nabla_{\mathbf{x}_{\mathrm{free}}} E(\theta,\mathbf{x}^*)=0$. The prediction of the model is then read from a subset of output vertices $\mathcal{O}\subset\mathcal{V}$ at convergence, that we denote  $\mathbf{x}^*_{\mathcal{O}}$. 

%This mirrors how training is defined in This is analogous to fixing the potentials of some nodes of a resistive network with voltage sources. This mirrors measuring the potential at other nodes of the circuit, or the displacement of other nodes of the elastic network, once the system has settled. 
%
%$\dot{\mathbf{x}}_{\mathrm{free}} = -\sfrac{\partial E}{\partial\mathbf{x}_{\mathrm{free}}}$. A free equilibrium $\mathbf{x}^0$ is 

While the inference rule intuitively follows from the idea of a dynamical system converging to an equilibrium, understanding how the parameters of such a model can be trained is less straightforward. Here, we will use the language of \emph{equilibrium propagation} (EP), a more general method developed to train arbitrary physical systems \citep{scellier2022agnostic}, already adapted to the energy function of predictive coding \citep{kerjan2026training}.  

\paragraph{Learning.} Once a free equilibrium $\mathbf{x}^*$ has been reached, EP can be used to estimate the gradient of the loss by contrasting it against a second weakly perturbed equilibrium, rather than by differentiating through the inference dynamics that produced it. Let $C(\mathbf{x}_{\mathcal{O}},\mathbf{y})$ be a loss function that compares the value nodes of the output vertices to the target $\mathbf{y}$, and let $\beta \in \mathbb{R}^+$ be a small constant. Let us assume we have already performed the inference phase over a given input $\mathbf{u}$, and that we have converged to an equilibrium state $\mathbf{x}^*$. The loss is then applied as a weak external force acting on the output vertices, giving the energy
$F_\beta(\theta,\mathbf{x},\mathbf{y})
    = E(\theta,\mathbf{x}) + \beta\, C(\mathbf{x}_{\mathcal{O}},\mathbf{y})$.
Keeping the input vertices clamped to $\mathbf{u}$ and letting the free value nodes relax on $F_\beta$, starting from $\mathbf{x}^*$, yields a second equilibrium point $\mathbf{x}^\beta$. EP then estimates $\nabla_\theta\mathcal{L}$ via contrastive learning as follows:
\begin{equation}
    G_{\mathrm{EP}}(\beta)
    = \frac{1}{\beta}\left[
        \frac{\partial E}{\partial\theta}(\theta,\mathbf{x}^\beta)
        - \frac{\partial E}{\partial\theta}(\theta,\mathbf{x}^*)
      \right].
    \label{eq:ep}
\end{equation}
Hence, the parameters are updated as $\theta\leftarrow\theta-\eta\,G_{\mathrm{EP}}(\beta)$, with learning rate $\eta$. In the case of feedforward architectures, the value nodes at the free equilibrium of a predictive coding model coincide with their predictions. This means that all error nodes vanish at the free state, so that $\frac{\partial E}{\partial\theta}(\theta,\mathbf{x}^*)=0$, and Eq.~\eqref{eq:ep} reduces to $G_{\mathrm{EP}}(\beta)=\frac{1}{\beta}\frac{\partial E}{\partial\theta}(\theta,\mathbf{x}^\beta)$ \citep{kerjan2026training}.

\paragraph{Nudging schemes.} The above equation is a finite-difference approximation of $\nabla_\theta\mathcal{L}$. Different choices of the sign of $\beta$ therefore result in different estimators of the same gradient. The simplest one, \emph{positive nudging}, uses a single $\beta>0$, which pulls the output vertices towards the target. This is a forward difference, and its bias with respect to the true gradient is of order $O(\beta)$. \emph{Centered nudging} \citep{laborieux2021scaling} removes this first-order term by relaxing to two equilibria, $\mathbf{x}^{\beta}$ and $\mathbf{x}^{-\beta}$, obtained by nudging the output vertices towards and away from the target, and contrasting them as follows:
\begin{equation}
    G_{\mathrm{EP}}^{\mathrm{c}}(\beta)
    = \frac{1}{2\beta}\left[
        \frac{\partial E}{\partial\theta}(\theta,\mathbf{x}^{\beta})
        - \frac{\partial E}{\partial\theta}(\theta,\mathbf{x}^{-\beta})
      \right],
    \label{eq:ep_centered}
\end{equation}
whose bias is of order $O(\beta^2)$, at the cost of an additional nudged phase. \emph{Random nudging} \citep{scellier17} keeps a single nudged phase, but samples the sign of $\beta$ uniformly at random for every data point. 

\subsection{Deep Equilibrium Models}
\label{sec:deq}

A deep equilibrium model \citep{bai19} replaces a stack of layers by a single map $f_\theta(\cdot,\mathbf{u}):\mathbb{R}^{n}\to\mathbb{R}^{n}$ applied to its own
output: given an input $\mathbf{u}$, the representation of the network is defined
implicitly as a fixed point $\mathbf{z}^*=f_\theta(\mathbf{z}^*,\mathbf{u})$, from which
the prediction is read by a decoder $d$, in this work always a linear classifier. Iterating $f_\theta(\cdot,\mathbf{u})$ from any
initial state converges to $\mathbf{z}^*$ whenever the map is contractive. Running $K$ iterations is equivalent to a forward pass through a weight-tied network of depth $K \times L$, where $L$ is the number of layers composing $f_\theta$. The number of iterations needed to reach the fixed point depends both on the architecture, through the contraction rate of $f_\theta$, and on the difficulty of the input.

\paragraph{Learning.}
As unrolling the $K$ iterations and applying backprop through time requires $O(K)$ memory, this method relies on the implicit function theorem to differentiate through the fixed point directly.
Writing $\ell(\mathbf{z},\theta)=C(d(\mathbf{z}),\mathbf{y})$ and
$\mathcal{L}(\theta)=\ell(\mathbf{z}^*,\theta)$, we have the following gradient of the loss:
\begin{equation}
    \nabla_\theta\mathcal{L}
    = \partial_\theta\ell
    + (\partial_\theta f_\theta)^{\!\top}
      \bigl(\mathbf{I}-\mathbf{J}_{f}(\mathbf{z}^*)^{\!\top}\bigr)^{-1}\nabla_{\mathbf{z}}\ell,
    \label{eq:ift}
\end{equation}
where all terms are evaluated at $\mathbf{z}^*$ and $\mathbf{J}_f$ is the Jacobian of
$f_\theta$ with respect to $\mathbf{z}$. The first term is the gradient with respect to
the decoder parameters. The linear solve with $(\mathbf{I}-\mathbf{J}_f^{\top})$ couples
every unit to every other, and requires $O(1)$ memory in the number of iterations. However, this linear quantity 
is global, and cannot be directly computed via local computations only. Luckily, we can rely on previous work that connects the implicit function theorem to EP to derive our results in the next sections \citep{zucchet2022beyond}.

\section{Methods}

Consider a PC graph with the hierarchical structure typical of a multilayer  network, depicted in Figure~\ref{fig:models_intro}(b): an ordered chain of hidden vertices $0,\ldots,L$, in which the prediction $\hat{\mathbf{x}}_i$ is computed from the value node $\mathbf{x}_{i-1}$ only. The input vertex is connected to the first hidden vertex via an embedding $e$, and the last hidden vertex is connected to the output vertex via a decoder $d$, that is, $\hat{\mathbf{x}}_0 = e(\mathbf{u})$ and $\hat{\mathbf{x}}_{\mathcal{O}} = d(\mathbf{x}_L)$. We call \emph{circular predictive coding network} (CPCN) the model obtained by closing this chain into a loop, that is, by identifying its first and last hidden vertices, so that $\mathbf{x}_0 \equiv \mathbf{x}_L$. The prediction $\hat{\mathbf{x}}_L$ is then the sum of the embedded input and of the prediction computed from $\mathbf{x}_{L-1}$. This model, depicted in Figure~\ref{fig:models_intro}(c), has the following local predictions:
\begin{equation}
    \hat{\mathbf{x}}_i = g_{i-1,i}(\mathbf{x}_{i-1}), \quad 1 \leq i < L, \qquad
    \hat{\mathbf{x}}_L = g_{L-1,L}(\mathbf{x}_{L-1}) + e(\mathbf{u}), \qquad
    \hat{\mathbf{x}}_{\mathcal{O}} = d(\mathbf{x}_L),
    \label{eq:cpc-pred}
\end{equation}
where indices are read modulo $L$, so that $\hat{\mathbf{x}}_1 = g_{L,1}(\mathbf{x}_L)$. Hence, the value node $\mathbf{x}_L$ plays three roles at once: it receives the input, it closes the loop, and it is read out by the decoder. Explicitly, this means that it has the following dynamics:
\begin{equation}
    \dot{\mathbf{x}}_L
    = -\frac{\partial E}{\partial \mathbf{x}_L}
    = -\boldsymbol{\varepsilon}_L
      + \mathbf{J}_{g_{L,1}}(\mathbf{x}_L)^{\!\top}\boldsymbol{\varepsilon}_1
      + \mathbf{J}_{d}(\mathbf{x}_L)^{\!\top}\boldsymbol{\varepsilon}_{\mathcal{O}},
    \label{eq:cpc-dyn}
\end{equation}
where $\mathbf{J}_g$ denotes the Jacobian of $g$, and $\boldsymbol{\varepsilon}_L = \mathbf{x}_L - g_{L-1,L}(\mathbf{x}_{L-1}) - e(\mathbf{u})$. The first term carries the stimulus, the second the error coming back around the loop, and the third the error at the output $\boldsymbol{\varepsilon}_{\mathcal{O}}$. Every other value node follows $\dot{\mathbf{x}}_i = -\boldsymbol{\varepsilon}_i + \mathbf{J}_{g_{i,i+1}}(\mathbf{x}_i)^{\!\top}\boldsymbol{\varepsilon}_{i+1}$, and hence only depends on its neighbours in the loop.

\paragraph{State of zero energy.}
The energy of Eq.~\eqref{eq:energy} vanishes if and only if every value node equals its prediction. Substituting the predictions of Eq.~\eqref{eq:cpc-pred} around the loop, this holds if and only if $\mathbf{x}_{\mathcal{O}}=d(\mathbf{x}_L)$ and
\begin{equation}
    \mathbf{x}_L = f_\theta(\mathbf{x}_L,\mathbf{u})
    := \bigl(g_{L-1,L}\circ\cdots\circ g_{1,2}\circ g_{L,1}\bigr)(\mathbf{x}_L) + e(\mathbf{u}),
    \label{eq:cpc-deq}
\end{equation}
which is the fixed point $\mathbf{z}^* = \mathbf{x}_L$ of a DEQ with transformation $f_\theta$ and decoder $d$. Unlike in feedforward models, this state cannot be computed by a single forward pass, and must be reached by relaxing the energy. In practice, we discretize the gradient flow, and call a \emph{sweep} one pass of updates around the loop. % CHECK: must match your implementation and the 3+2t count
As $K$ iterations of $f_\theta$ correspond to a weight-tied network of depth $K\times L$, the effective depth of a CPCN is related to how long the energy relaxes, not by the number of vertices in the loop. Finally, as in the feedforward case, all error nodes vanish at the free equilibrium, so $\frac{\partial E}{\partial\theta}(\theta,\mathbf{x}^*)=0$ and the simplified form of Eq.~\eqref{eq:ep} still applies.

\paragraph{Dual readout.}
In Eq.~\eqref{eq:cpc-pred}, the decoder reads $\mathbf{x}_L$ only, so the nudge on the output vertex reaches the rest of the loop only after travelling around it. In the \emph{dual-readout} variant, the decoder instead reads the average of the last two value nodes of the loop, $\hat{\mathbf{x}}_{\mathcal{O}} = d\bigl(\tfrac{1}{2}(\mathbf{x}_{L-1}+\mathbf{x}_L)\bigr)$, which adds no parameters. 
As $\mathbf{x}_{\mathcal{O}}$ is free, its error vanishes at any free equilibrium, so the loop reaches the same zero-energy state as in the single-readout model. Only the nudged phase changes: the output error now enters the loop at two value nodes instead of one, and for $L=2$ reaches all of them directly. We will test this variation on computer vision tasks, showing that it outperforms the single readout models by a significant margin. A sketch of this model is in Fig.~\ref{fig:cifar}(left). 

\begin{algorithm}[t]
\caption{Training a recursive CPCN on one batch.}
\label{alg:recursive}
\begin{algorithmic}[1]
\Require puzzles $\mathbf{u}$, targets $\mathbf{y}$, number of stages $S$
\State answer $\mathbf{a} \gets \mathbf{0}$, \ state $\mathbf{x} \gets \mathbf{0}$
\For{$s = 1,\ldots,S$}
    \State $\mathbf{x}^* \gets$ free phase with inputs $(\mathbf{u},\mathbf{a})$, starting from $\mathbf{x}$
    \State $\mathbf{x}^{\beta} \gets$ nudged phase towards $\mathbf{y}$, starting from $\mathbf{x}^*$
    \State optimizer step on $G_{\mathrm{EP}}$ \Comment{local update, $\mathbf{a}$ held constant}
    \State $\mathbf{a} \gets \mathbf{x}^*_L$, \ $\mathbf{x} \gets \mathbf{x}^*$ \Comment{carry answer and state}
\EndFor
\end{algorithmic}
\end{algorithm}

\subsection{Recursive CPCNs}
\label{sec:recursive}

Recursive reasoning models, such as HRM \citep{wang2025hierarchicalreasoningmodel} and TRM \citep{jolicoeur2025less}, nest several recurrent loops, supervise every outer iteration, and are highly parameter-efficient on constraint satisfaction tasks such as Sudoku. We extend CPCNs in the same spirit, and call the result \emph{recursive CPCN}: we chain a sequence of energy minimizations, which we call \emph{stages}, all sharing the same weights. At every stage, the readout node also receives the \emph{answer state} $\mathbf{a}$, i.e., the free equilibrium $\mathbf{x}^*_L$ of the previous stage, through a map $\mathbf{W}_a$:
\begin{equation}
    \hat{\mathbf{x}}_L = g_{L-1,L}(\mathbf{x}_{L-1}) + e(\mathbf{u}) + \mathbf{W}_a\mathbf{a}.
\end{equation}
At the first stage, $\mathbf{a}=\mathbf{0}$ and the model reduces to a standard CPCN. Intuitively, a standard CPCN must solve the whole puzzle within a single relaxation, while a recursive one only needs to improve on its previous answer at every stage. Hard puzzles can then be solved progressively, in the same way a person fills in a Sudoku in rounds, using the cells found so far to deduce the next ones.

Since the answer state is held fixed while a stage relaxes, it is simply one more clamped input, and the results of Section~\ref{sec:theory} apply to every stage separately. Each stage is warm-started from the free state of the previous one, and is followed by its own nudged phase, EP updates, and optimizer steps (Algorithm~\ref{alg:recursive}). As every stage is supervised, the same weights learn both to propose an answer, when no previous answer is available, and to correct one. During training, the number of stages is sampled uniformly between $4$ and $16$ for every batch; from the fifth stage on, the nudged phase, and hence the update, only uses half of the batch, three quarters of which are samples that the previous stage did not solve. We test this variant on Sudoku in Section~\ref{sec:sudoku}, and show that a recursive CPCN outperforms a standard CPCN one despite being $10\times$ smaller.

\section{Theoretical Results}
\label{sec:theory}

\begin{wrapfigure}{r}
{0.3\textwidth}
\vspace{-1.5cm}
    \centering
    \includegraphics[width=\linewidth]{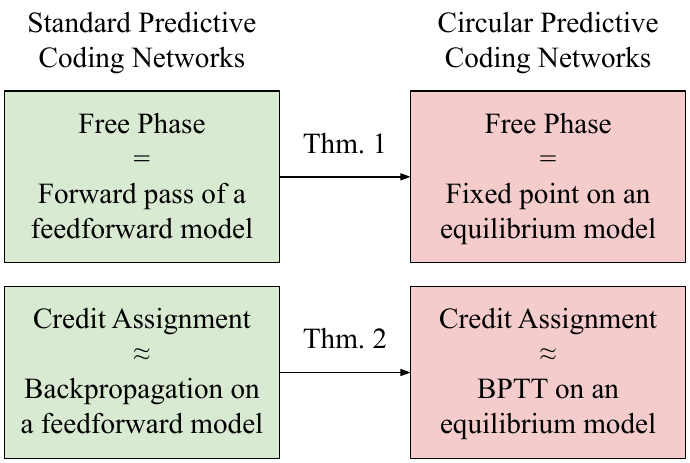}
    \caption{Meaning of our results.}
    \label{fig:sketch}
    \vspace{-3ex}
\end{wrapfigure}

The two properties that make feedforward PC models easy to train and develop are (i) the relaxation process of the free phase converges to a state of zero energy, that is equivalent to the one reached by a standard forward pass; (ii) the weight update is able to approximate that of backprop \citep{whittington2017approximation,millidge2020predictive}. In this section, we formally show that these two facts have their conceptual equivalents in circular PC models and DEQs (Fig.~\ref{fig:sketch}). Every theorem in this section is presented informally for clarity of exposition; we refer to the supplementary material for the formal statements and proofs, and for empirical validations of our theoretical results, where we show that  the described behavior is also observed
in practice under numerical simulations.

Let us assume we have a circular PC model $M$ with edge maps $g_{L,1},g_{1,2},\ldots,g_{L-1,L}$, embedding $e$ and decoder $d$, as in Eq.~\eqref{eq:cpc-pred}, and let $M'$ be the DEQ built from those same maps, whose implicit layer is their composition around the loop $f_\theta(\cdot,\mathbf{u})$, and whose output is read via $d$. The two models are structurally identical: same layers and same parameters. They differ only in how they perform inference: $M$ relaxes all of its value nodes along the gradient flow of the energy, while $M'$ iterates $f_\theta(\cdot,\mathbf{u})$ until convergence via a fixed point iterator. 

If every parametric map of the loop is a contraction, so is their composition, and hence the zero-energy state exists and is unique. Furthermore, at any stationary point of the energy we have $\boldsymbol{\varepsilon}_{\mathcal{O}}=0$, and Eq.~\eqref{eq:cpc-dyn}, together with the dynamics of the other value nodes, gives $\boldsymbol{\varepsilon}_i = \mathbf{J}_{g_{i,i+1}}(\mathbf{x}_i)^{\!\top}\boldsymbol{\varepsilon}_{i+1}$ for every $i$. This forces every error node to vanish, and hence no stationary point other than the zero-energy one can exist. We hence have the following:

\begin{theorem}[Inference]
\label{thm:inference}
Let every map of the loop be $C^1$ and $q$-Lipschitz in the Euclidean norm, with $q<1$. Then:
\begin{enumerate}[label=(\roman*), nosep]
  \item the energy of $M$ has a unique stationary point $\mathbf{x}^*$ in the free value
        nodes, which is its global minimum and has zero energy;
  \item $\mathbf{x}^*_L$ is the unique fixed point of $f_\theta(\cdot,\mathbf{u})$, and
        every value node of $\mathbf{x}^*$, including the output $\mathbf{x}^*_{\mathcal{O}}$,
        equals the corresponding latent variable of $M'$;
  \item if, in addition, the maps of the loop and the decoder are $C^2$, the gradient flow
        of the energy converges to $\mathbf{x}^*$ from any initialization.
\end{enumerate}
\end{theorem}

The free phase of a circular model and the forward pass of $M'$ therefore return the same output, and the same latent variables: the relaxation reaches, value node by value node, the state that iterating the composed map reaches, and it cannot stall anywhere else. This is the counterpart, for circular models, of the fact that the free phase of a feedforward PC model converges to a state of zero energy, and produces the same output and hidden states as a forward pass. To then show a connection between how parameters are updated in the two methods, we need to rely on previous results, which have studied the relations between EP and the implicit function theorem \citep{scellier17,zucchet2022beyond}.%\footnote{Note that, as the error nodes vanish at $\mathbf{x}^*$, the update of Eq.~\eqref{eq:ep} is still the weight update of PC, computed at the nudged equilibrium. What differs from the standard training of PC models is that the output is weakly nudged towards the target, rather than clamped to it: Theorem~\ref{thm:learning} holds in the limit $\beta\to0$, and would have not held had the output been clamped.}

The intuition behind the second result is the following. The nudged equilibrium satisfies $\nabla_{\mathbf{x}}E(\theta,\mathbf{x}^\beta) + \beta\nabla_{\mathbf{x}}C = 0$. Differentiating this condition with respect to $\beta$ shows that, to first order, the nudge moves the free value nodes by $-\beta\mathbf{H}^{-1}\nabla_{\mathbf{x}}C$, where $\mathbf{H}$ is the Hessian of $E$ in the free value nodes, which is positive definite at $\mathbf{x}^*$ under the assumptions of Theorem~\ref{thm:inference}. This is the same linear response that the implicit function theorem uses to differentiate $\mathbf{x}^*$ with respect to $\theta$. Hence, the contrast of Eq.~\eqref{eq:ep} converges to $-\frac{\partial^2 E}{\partial\theta\,\partial\mathbf{x}}\mathbf{H}^{-1}\nabla_{\mathbf{x}}C=\nabla_\theta\mathcal{L}$: EP computes the implicit gradient by letting the value nodes relax, rather than by inverting $\mathbf{H}$. More formally, we have the following, inspired and adapted from existing work \citep{zucchet2022beyond,scellier17}:

\begin{theorem}[Learning]
\label{thm:learning}
Let the assumptions of Theorem~\ref{thm:inference} hold, let every map of $M$ be twice continuously differentiable jointly in the parameters and the value nodes, and let $C$ be twice continuously differentiable. Write $\mathcal{L}(\theta)=C(\mathbf{x}^*_{\mathcal{O}}(\theta),\mathbf{y})$ for the cost at the free equilibrium. If both phases are solved exactly, then $G_{\mathrm{EP}}(\beta)\to\nabla_\theta\mathcal{L}(\theta)$ as $\beta\to0$. % and, given the smoothness required by the corresponding Taylor expansions, the error is $O(\beta)$, or $O(\beta^2)$ for the centered estimator of Eq.~\eqref{eq:ep_centered}.
\end{theorem}

By Theorem~\ref{thm:inference}, $\mathbf{x}^*_{\mathcal{O}}$ is the output of $M'$, and hence $\nabla_\theta\mathcal{L}$ is exactly the implicit gradient of the DEQ: training a circular PC model with EP approximates the training of its equivalent DEQ via the implicit function theorem. This is the counterpart, for circular models, of the fact that the weight update of feedforward PC models approximates that of backprop.

\section{Experiments}

We now evaluate the performance of circular predictive coding on three tasks: image classification, language modeling, and constraint satisfaction problems. The datasets we use are standard in the literature: CIFAR10/100, and Imagenet32 for computer vision, Tiny Stories \citep{eldan2023tinystories}, and Wikitext-103 \citep{merity2017wiki103} for natural language modeling, and sudoku-hard for constraint satisfaction \citep{palm2018recurrent}.

\subsection{Image Classification}

\begin{figure}[t]
\centering
\includegraphics[width=0.8\linewidth]{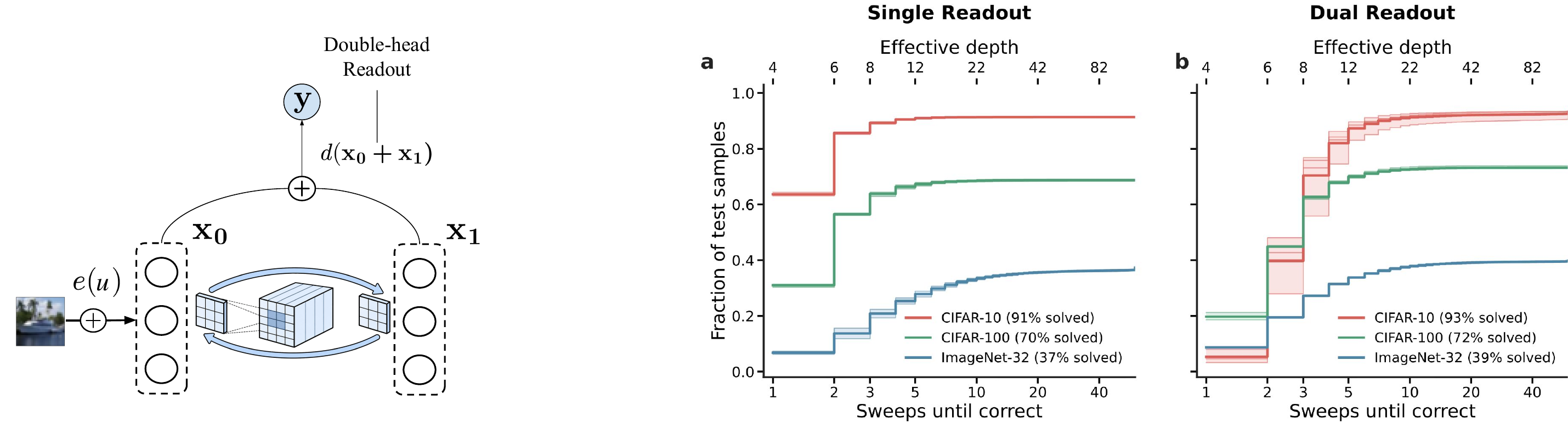}
%\vspace{-0.5cm}
\caption{Left: A sketch of the dual-readout model, where both value nodes influence the decoder and, consequently, are directly influenced by the nudging signal. Right: Fraction of test images that are classified correctly and stay correct from sweep $t$ onward, for CIFAR-10, CIFAR-100 and ImageNet32. (a) Single readout (the head reads one value node). (b) Dual readout (the head reads the average of the two value nodes). The top axis shows the effective depth $3+2t$. 
}
\label{fig:cifar}
%\vspace{-0.5cm}
\end{figure}

\begin{table}[t]
\centering
\caption{Accuracy on CIFAR-10/100. The PC experiments with a similar number of parameters have been run by us using the official repo \citep{pinchetti2025benchmarking}, EP stands for the EP formulation of PC \citep{kerjan2026training}; the ones on  ResNet-18 and VGG-15 are from the original work \citep{qi2025towards}.}
\label{tab:cifar_accuracy}
\scriptsize
\setlength{\tabcolsep}{2.2pt}
\begin{tabular}{lcc|cccc|cc|c|c}
\toprule
& \multicolumn{2}{c|}{}
& \multicolumn{4}{c|}{VGG5 - Similar number of parameters}
& \multicolumn{2}{c|}{ResNet-18} 
& \multicolumn{1}{c|}{VGG-15}
& \multicolumn{1}{c}{} \\
\cmidrule(lr){2-7}\cmidrule(lr){8-10}
Dataset
& CPCN & DualCPCN
&  PC & PC-CN & iPC & EP
& PC + FU & iPC + S & iPC + S & DEQ \\
\midrule
CIFAR-10
& $91.42{\scriptstyle\,\pm\,0.71}$
& $\mathbf{93.54}{\scriptstyle\,\pm\,0.11}$
& $86.93{\scriptstyle\,\pm\,0.29}$
& $88.41{\scriptstyle\,\pm\,0.18}$
& $85.62{\scriptstyle\,\pm\,0.35}$
& $90.01{\scriptstyle\,\pm\,0.21}$
& $92.39{\scriptstyle\,\pm\,0.04}$
& $92.47{\scriptstyle\,\pm\,0.01}$
& -
& $91.84{\scriptstyle\,\pm\,0.09}$ \\

CIFAR-100
& $70.48{\scriptstyle\,\pm\,0.12}$
& $\mathbf{72.38}{\scriptstyle\,\pm\,0.11}$
& $62.91{\scriptstyle\,\pm\,0.73}$
& $63.99{\scriptstyle\,\pm\,0.11}$
& $61.48{\scriptstyle\,\pm\,0.58}$
& $68.42{\scriptstyle\,\pm\,0.26}$
& $70.38{\scriptstyle\,\pm\,0.20}$
& $70.90{\scriptstyle\,\pm\,0.18}$
& -
& $70.81{\scriptstyle\,\pm\,0.07}$ \\

% CIFAR-100 (top5)
% & $87.22{\scriptstyle\,\pm\,0.18}$
% & $\mathbf{92.66}{\scriptstyle\,\pm\,0.10}$
% & $87.14{\scriptstyle\,\pm\,0.25}$
% & $82.13{\scriptstyle\,\pm\,0.51}$
% & $84.11{\scriptstyle\,\pm\,0.16}$
% & $80.56{\scriptstyle\,\pm\,0.43}$
% & $90.74{\scriptstyle\,\pm\,0.10}$
% & $90.47{\scriptstyle\,\pm\,0.11}$ \\

ImageNet32
& $37.43{\scriptstyle\,\pm\,0.24}$
& $\mathbf{39.31}{\scriptstyle\,\pm\,0.24}$
& $31.23{\scriptstyle\,\pm\,1.68}$
& $33.15{\scriptstyle\,\pm\,1.20}$
& $30.35{\scriptstyle\,\pm\,1.73}$
& $34.11{\scriptstyle\,\pm\,1.54}$
& $-$
& $-$ 
& $36.23$
& $39.15{\scriptstyle\,\pm\,0.10}$ \\

\bottomrule
\end{tabular}
\label{tab:cifar}
\end{table}

On CIFAR-10/100, we use a CPCN with a loop that has two layers, each a $256$-channel $8\times8$ feature map, connected by two convolutional layers. The embedding $e$ is a three-layer convolutional encoder still trained with predictive coding, and the decoder $d$ is applied after global pooling. One sweep on the trained model therefore corresponds to a network of depth $5$, and $t$ sweeps to an effective depth of $3+2t$. On ImageNet32, we double all widths, which gives $7.2$M parameters against $1.5$M on CIFAR. We train for $180$ epochs on CIFAR-10/100 and $50$ on ImageNet32. We compare against PC models with a VGG-5 architecture and a comparable number of parameters, against state-of-the-art PC variants on ResNet-18 and VGG-15 \citep{qi2025towards}, and against DEQ models with the same architecture and training recipe as the EP runs, and differ only in the gradient: the exact implicit-function-theorem gradient, with the adjoint solved. As data augmentation, we use random crops and horizontal flips, the same as the work we compare against.

\paragraph{Results.}
Dual-readout CPCNs reach the highest accuracy on all three datasets (Table~\ref{tab:cifar}), above both the VGG-5 baselines of comparable size and the much larger ResNet-18 and VGG-15 models. Single-readout CPCNs outperform all VGG-5 baselines on CIFAR-10/100 and ImageNet32. The dual readout thus adds $2.12$, $1.90$, and $1.88$ points. We attribute this gain to the architecture rather than to the learning rule: reading $\mathbf{x}_{L-1}$ gives the cost a direct path to it, as a skip connection would, whereas with a single readout the error reaching $\mathbf{x}_{L-1}$ must first cross $g_{L-1,L}$.

\paragraph{Convergence.}
Figure~\ref{fig:cifar}(right) shows when each test image becomes, and stays, correctly classified. The single readout is accurate early: after one sweep, it already classifies $64\%$ of CIFAR-10 and $31\%$ of CIFAR-100 images correctly, against $5\%$ and $20\%$ for the dual readout, which catches up after about three sweeps and keeps improving until sweeps $5$ to $10$. ImageNet32 converges more slowly under both readouts, with $28\%$ (single) and $34\%$ (dual) of images correct at sweep $5$, and levels off around sweep $20$: harder datasets need more sweeps, that is, more effective depth.

\begin{figure}[t]
\centering
\includegraphics[width=\linewidth]{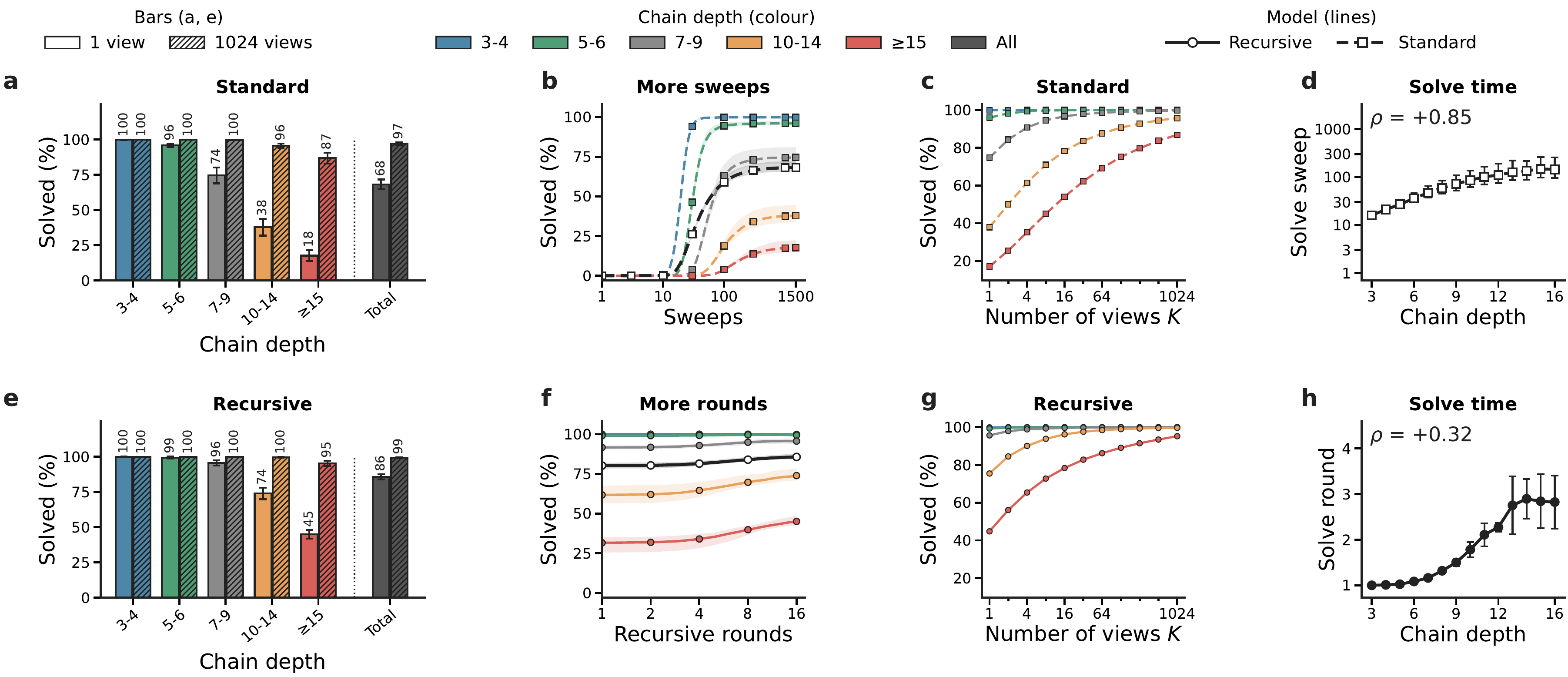}
%\vspace{-0.5cm}
\caption{Top row: the standard model; bottom row: the recursive model (16 recursive stages). Boards are grouped by chain depth, which is the number of dependent deduction steps needed to solve the puzzle. (a, e) Board accuracy per chain-depth group and in total, for a single view (open bars) and the lowest self-cost of 1024 digit-relabelled views. (b, f) Accuracy as the number of sweeps or recursive rounds grows. (c, g) Accuracy as the number of views increases from $1$ to $1024$. (d, h) The round at which a board becomes correct and stays correct, against its chain depth; $\rho$ is the Spearman correlation between the two. 
}
\label{fig:sudoku}
%\vspace{-0.5cm}
\end{figure}

\subsection{Sudoku}
\label{sec:sudoku}

We now test whether the relaxation of a CPCN can carry out multi-step reasoning, comparing a standard CPCN with the recursive CPCN of Section~\ref{sec:recursive}.
We use Sudoku-hard \citep{palm2018recurrent}, with $180$k training, $18$k validation and $18$k test puzzles, each with $17$ to $34$ given cells. We grade each test board by its \emph{chain depth}: the number of rounds a reference solver needs to complete it, where each round fills every cell forced by a naked single (only one digit fits the cell) or a hidden single (the cell is the only place for a digit in its row, column or box), and reveals one true digit if no cell is forced. The chain depth thus approximates the length of the longest chain of dependent deductions.

\paragraph{Setup.}
The loop is an MLP-Mixer block \citep{tolstikhin2021mlp} with $L=4$ value nodes, one per layer of its token-mixing MLP, which exchanges information across the $81$ cells, and of its channel-mixing MLP, which updates each cell. The standard model has width $512$ ($1.1$M parameters) and runs for $1500$ sweeps; the recursive model has width $120$ ($105$k parameters) and runs for $32$ stages at test time. A board is solved only if all $81$ cells are correct. In the \emph{single-view} evaluation, each board is solved once, from a zero state. In the \emph{multi-view} evaluation, we relabel its digits with $1024$ fixed permutations, solve each view independently, map the predictions back, and keep the one with the lowest \emph{self-cost}, i.e., the cross-entropy between the output $\mathbf{x}_{\mathcal{O}}$ and its own most likely digits. This selection uses neither labels nor the rules of Sudoku. Full details are in Appendix~\ref{app:sudoku}.

\paragraph{Results.}
Despite being ten times smaller, the recursive model is more accurate at every chain depth (Fig.~\ref{fig:sudoku}(a, e)): it solves $85.7\%$ of the boards against $68.2\%$ with one view, and $99.3\%$ against $97.2\%$ with $1024$ views. The gap grows with depth: on boards of depth $\geq 15$, it is $45\%$ against $18\%$ with one view, and $95\%$ against $87\%$ with $1024$ views. Multiple views help both models most on deep boards, and the self-cost is a reliable selector, within $0.2$ points of an oracle that picks any correct view. The two models use test-time compute differently. The standard model is still improving on hard boards after $500$ sweeps (b), and the sweep at which a board is solved grows with its chain depth ($\rho=0.85$, d), as if the relaxation carried out the deductions one after the other. The recursive model solves $80\%$ of the boards at its first stage, and improves on hard ones up to about $16$ stages; (f). Its solve stage depends only weakly on depth ($\rho=0.32$, h): only the most complex boards benefit from this, and by a margin of about $10\%$.

\section{Language Modeling}
\label{sec:lm}

Transformer models are the dominant architecture of modern natural language modeling, yet they remain almost absent from the predictive coding literature: existing results are limited to small feedforward models trained on small language-modeling datasets \citep{pinchetti2022predictive, salvatori2024predictive}. We take a step towards closing this gap, and train CPCNs whose loop is a single autoregressive transformer block \citep{vaswani2017attention} on TinyStories \citep{eldan2023tinystories} and on WikiText-103 \citep{merity2017wiki103}.

\textbf{Setup.} The loop is one causal transformer block, self-attention followed by a two-layer MLP, with $L=3$ value nodes: after the attention sub-layer, at the hidden layer of the MLP, and at its output, which feeds the linear readout and the attention of the next sweep. The same model is used on both datasets: text is tokenized with a BPE vocabulary of $8192$ tokens trained on each dataset, and the model has width $384$, $6$ heads, a context of $128$ tokens, and $8.7$M parameters. Every matrix of the loop has its spectral norm capped at $1$; we also bound the gain of the LayerNorm inside the loop, and normalize queries and keys so that attention logits remain bounded. Both training sets have $48.1$M tokens: a slice of TinyStories, and the corresponding prefix of the WikiText-103 training split. As a baseline, we train the same recurrent model with the implicit gradient of its fixed point (DEQ). All models are trained for $20$k steps with the same optimizer, over three seeds.

\textbf{Nudged phase.} Relaxing the nudged energy $F_\beta$ with the same sweeps as the free phase becomes slow on transformer blocks, so we reach the nudged equilibrium with a second-order method instead: starting from the free equilibrium, where the gradient of $F_\beta$ reduces to that of the cost, we take a single Newton step on $F_\beta$, using the curvature of the energy at that point. The step is computed with $T$ iterations of a momentum method \citep{polyak1964some} that only needs Hessian--vector products ($T = 150$ on TinyStories, $400$ on WikiText-103). The step size of this inner solver is adapted at every training step from a running estimate of the curvature of the energy. The parameters are then updated using centered nudging.

\begin{figure}[t]
  \centering
  \includegraphics[width=0.6\linewidth]{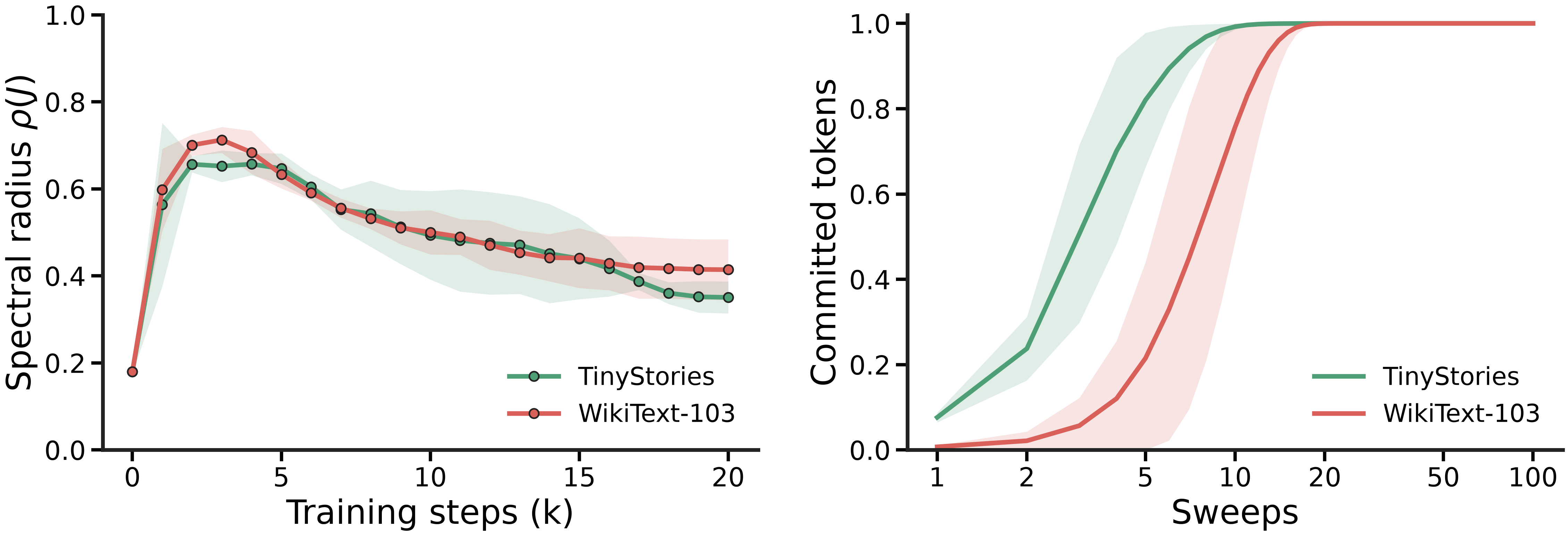}
    \caption{Use of the loop by CPC transformers trained with EP, on TinyStories (green) and WikiText-103 (red); mean $\pm$ std over three runs. Left: spectral radius $\rho(J)$ of the loop Jacobian at the free equilibrium during training. Right: fraction of validation tokens whose prediction has committed after $t$ sweeps from the zero state, that is, whose loss stays within $0.01$ nats of its converged value from sweep $t$ onward.}
  \label{fig:lm_depth}
\end{figure}

\textbf{Results.} Our CPCN matches its DEQ counterpart on both datasets (Table~\ref{tab:nlp}): it reaches a validation perplexity of $12.96 \pm 0.64$ against $13.33 \pm 0.39$ on TinyStories, and a test perplexity of $110.83 \pm 0.53$ against $112.59 \pm 0.85$ on WikiText-103, where it is slightly better on every seed. On both datasets, the spectral radius of the loop Jacobian peaks at about $0.7$ early in training and then decreases, to $0.35$ on TinyStories and $0.41 \pm 0.07$ on WikiText-103 (Figure~\ref{fig:lm_depth}). On WikiText-103 the median token commits to its final prediction after $6$ to $11$ sweeps, $95\%$ of them after $9$ to $16$; on TinyStories, the median token commits after $3$ to $4$ sweeps.

%\begin{table}[t]
%\caption{Perplexity ($\downarrow$) of the CPC transformer and of the same
%recurrent model trained with implicit-function-theorem gradients at its fixed point (DEQ), mean $\pm$ std over 3 seeds,
%20k training steps. Tokens: size of the training set. TinyStories: validation perplexity; LM1B: test perplexity.}
%\label{tab:nlp}
%\centering
%%\begin{tabular}{lccccc}
%\toprule
%Dataset & Training tokens & Parameters & CPCN & DEQ & PC \citep{pinchetti2022predictive}  \\
%\midrule
%TinyStories & 48.1M & 8.71M & $\mathbf{12.96\pm0.64}$ & $13.33\pm0.39$ & Diverged \\
%LM1B & 5.4M & 2.26M & $\mathbf{145.84\pm1.03}$ & $146.03\pm0.48$ & $\approx700$ \\
%\bottomrule
%\end{tabular}
%\end{table}

\begin{table}[t]
\caption{Perplexity ($\downarrow$) of the CPC transformer and of the same
recurrent model trained with implicit-function-theorem gradients at its fixed point (DEQ), mean $\pm$ std over 3 seeds,
20k training steps. Tokens: size of the training set (for WikiText-103, a prefix of the training split of the same size
as the TinyStories slice). TinyStories: validation perplexity; WikiText-103: test perplexity on the full official test split.}
\label{tab:nlp}
\centering
\begin{tabular}{lcccc}
\toprule
Dataset & Training tokens & Parameters & CPCN & DEQ \\
\midrule
TinyStories & 48.1M & 8.71M & $\mathbf{12.96\pm0.64}$ & $13.33\pm0.39$ \\
WikiText-103 & 48.1M & 8.71M & $\mathbf{110.83\pm0.53}$ & $112.59\pm0.85$ \\
\bottomrule
\end{tabular}
\end{table}
%\section{Language Modeling}
%\label{sec:lm}

\section{Related Works}

While predictive coding is a model of cortical processing \citep{rao1999predictive,friston2005theory}, it can be seen as a learning algorithm that approximates backpropagation \citep{whittington2017approximation, millidge2022predictive}, or matches it exactly under suitable schedules \citep{salvatori2022reverse}. Equilibrium propagation plays the same role for recurrent energy-based models \citep{scellier17}: it contrasts a free and a nudged equilibrium, and as the nudge vanishes its updates equal those of recurrent BP \citep{scellier2019equivalence}, match BP through time (BPTT) step by step for static inputs \citep{ernoult2019updates}, and compute an implicit gradient \citep{zucchet2022beyond}. Cycles in PC graphs \citep{salvatori2022learning} have been used for other ends \citep{oliviers2026bidirectional,salvatori2024predictive,seely2025sheaf}, but not for depth. In a CPCN, they make PC a recurrent energy-based model, so Theorem~2 transfers these results: EP on a CPCN approximates the implicit gradient of the equivalent DEQ, as PC on a feedforward network approximates BP. %Temporal PC also adds recurrence \citep{millidge2024predictive,potter2026learning}, but over input time steps rather than relaxation time.

\section{Conclusion}

We showed that a cyclic energy-based network can realize the equilibrium computation of an equilibrium model without differentiating through its relaxation dynamics. This yields a model that uses relaxation time as a computational resource on Sudoku, while remaining competitive on image classification tasks and language modeling. The practical interest of the construction lies in physical implementations: as depth is
unrolled in time rather than in space, a fixed-size analog loop could in principle spend more time settling on harder inputs using a fixed-size array, while learning reduces to a local contrast between two measured steady states \citep{kendall2020training,yi2023activity}.

\newpage

\bibliography{references}
\bibliographystyle{plainnat}

\newpage

\appendix

\appendix

\section{Empirical validation of the theory}
\label{sec:validation}

\begin{figure}[t]
\centering
\includegraphics[width=\linewidth]{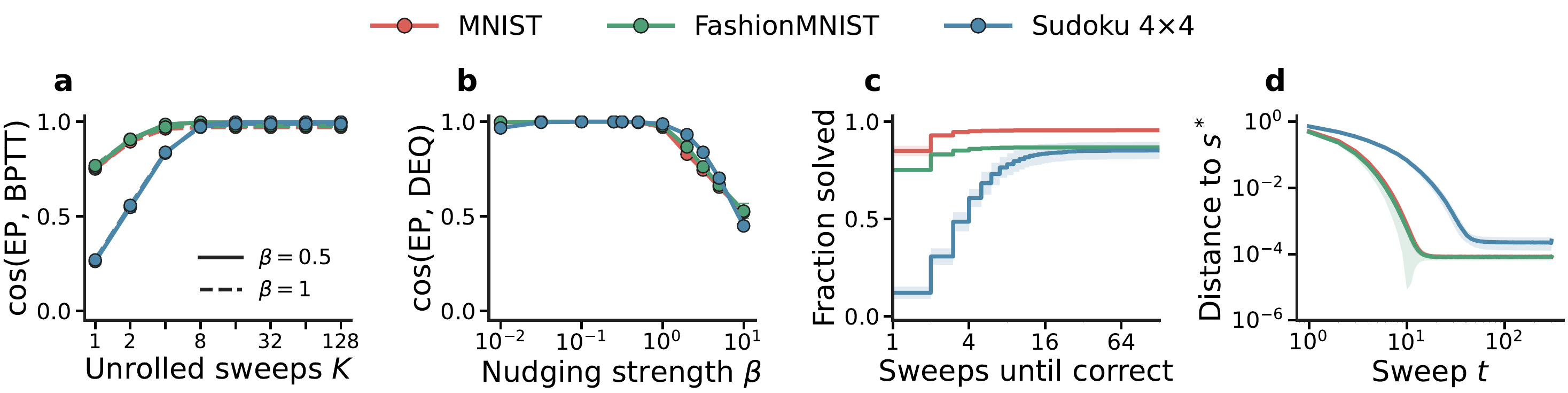}
%\vspace{-0.5cm}
\caption{(a) Cosine similarity between the EP weight gradient and the gradient of a $K$-sweep BPTT unroll, on $128$ samples. (b) Same EP gradients against the exact implicit-function gradient of a DEQ. Fidelity decays for large $\beta$, where the nudge moves the equilibrium beyond the linear regime EP relies on. (c) The sweep after which the loop's prediction is correct and never changes again. MNIST and FashionMNIST commit on the first sweep for almost every image ($95$th percentile: $2$ sweeps); Sudoku boards need a median of $3$ sweeps, and the hardest $5\%$ take $11$ or more. (d) The loop starts from the zero state and, after each sweep, we record the distance to the state it finally settles in, relative to the norm of that state (median over $1000$ test samples). On MNIST and FashionMNIST, the model converges in about $10$ to $20$ sweeps; Sudoku, less contractive, needs about $40$ to $60$. 
}
\label{fig:poc}
%\vspace{-0.5cm}
\end{figure}

Theorems~\ref{thm:inference} and~\ref{thm:learning} assume that the equilibria and fixed points are reached exactly, and that $\beta \to 0$. As neither assumption holds in numerical simulations, we now check empirically whether the predicted behavior is observed in practice. First, we check whether the EP update is a faithful estimate of both the BPTT and the IFT gradients; then, we show that some test examples need more sweeps than others to converge to the correct solution, hence proving that the circular model exhibits effective depth; finally, we check whether the relaxation reaches a unique equilibrium. As datasets, we consider MNIST, FashionMNIST, and $4 \times 4$ Sudokus. 

\paragraph{Gradient results.} We initialize a CPCN with a single embedding layer and two hidden value nodes ($L=2$) of width $128$, and, from the same weights, we compute the gradients given by BPTT, by EP with centered nudging, and by the IFT. The results, reported in Figure~\ref{fig:poc}(a,b), show that (i) BPTT matches the EP gradient only once it unrolls enough sweeps to reach the fixed point, and (ii) EP closely approximates the IFT gradient, with smaller values of $\beta$ giving better approximations.

\paragraph{Effective depth.}
Since the free equilibrium is unique (Theorem~\ref{thm:inference}), any procedure that drives the prediction errors to zero reaches the same state. To study the effective depth, we train a CPCN, and evaluate it as if it were a DEQ: via multiple \emph{sweeps} through the loop, where a sweep is a single application of $f_\theta(\cdot,\mathbf{u})$. 
We measure depth in sweeps rather than in gradient-flow steps because a sweep has a direct architectural meaning: it sets every value node to its prediction once around the loop, $\mathbf{x}_i \leftarrow \hat{\mathbf{x}}_i$, and hence applies each of the $L$ maps exactly once, whereas the number of gradient-flow steps depends on the integration step size. Since this update is local and reaches the same unique equilibrium as the gradient flow, the sweep count is a fair proxy for effective depth.

Given a test data point that is correctly predicted by the trained CPCN, its effective depth is $K \times L$, where $L$ is the number of vertices in the loop and $K$ the number of sweeps needed to converge to the correct solution. In Figure~\ref{fig:poc}(c), we observe that on MNIST and FashionMNIST the median input commits after one sweep, but harder-to-classify inputs need more than $4$ sweeps. On Sudoku, the median board needs $3$ sweeps, and the hardest $5\%$ need $11$ or more.

\paragraph{Convergence to an equilibrium.}
As all of the trained models in the figure have spectral norm $\|\mathbf{J}_f\|_2 < 1$, and converge to a unique state. (Figure~\ref{fig:poc}(d)): the relative distance to the fixed point decays on every task until convergence. The vision models settle at $\|\mathbf{J}_f\|_2 \approx 0.5$ and reach $10^{-4}$ in about $10$ sweeps, while Sudoku trains to $\|\mathbf{J}_f\|_2 = 0.85$ and needs about $40$ sweeps to reach convergence. % CHECK: caption says 20 and 60

\subsection{Theory as a guide, not a constraint}

Requiring every map in the loop to be a contraction is a strong condition: It guarantees that the free phase always converges to a single equilibrium, from any starting point. Training, however, only needs something weaker: that the free phase reaches a zero-energy state, and that this state is not degenerate. Whenever that happens, our results still apply locally. The state is the fixed point of the equivalent DEQ, small perturbations relax back to it, and EP estimates the implicit gradient there. Since strict contractivity limits what the model can express, we relax it in our larger experiments. Instead, we check directly that these local conditions hold, by measuring the residual at the end of the free phase, the spectral radius of the loop at equilibrium, and the agreement between EP and implicit gradients. The price is that the equilibrium may no longer be unique, so the state the model reaches depends on where relaxation starts, which we always fix at zero. Interestingly, energy relaxation stays stable even in cases where simply iterating the DEQ would oscillate. 

We use this view to design our experiments. The theory tells us what must be preserved: the free phase should reach a zero-energy state, that state should be stable, and the nudge should stay small enough for EP to track the implicit gradient. It also tells us what is free: only the final equilibrium matters, not how it is reached. Within these constraints, we choose whatever gives the best performance. We bound the gain of the loop rather than forcing it to contract. We reach equilibrium with faster solvers rather than plain relaxation. And we shape the nudged phase so that the learning signal reaches the loop more easily.

\section{Image classification: experimental details}
\label{app:vision}

\paragraph{Data.} CIFAR-10 and CIFAR-100 have 50{,}000 training and 10{,}000 test images. For ImageNet32 we use the $32\times32$ version of ILSVRC-2012, with 1{,}281{,}167 training images, and its 50{,}000 validation images as the test set. Images are normalised per channel with training-set statistics. The only augmentations are a random $32\times32$ crop after reflection-padding by 4 pixels and a horizontal flip with probability $1/2$.

\paragraph{Architecture.} The encoder $e$ has three convolutions (Table~\ref{tab:vision-arch}). The first two are each followed by a ReLU and an affine group norm with 8 groups. The third is linear and gives the embedded input $e(\mathbf{u})\in\mathbb{R}^{8\times8\times w}$ of the loop. Each layer of the encoder has its own value node that keeps information processing local. The loop has two value nodes $\mathbf{x}_1,\mathbf{x}_2\in\mathbb{R}^{8\times8\times w}$ and two $3\times3$ convolutions:
\begin{equation}
\hat{\mathbf{x}}_1=\sigma\big(\mathbf{W}_1*\mathbf{x}_2+\mathbf{b}_1\big),\qquad
\hat{\mathbf{x}}_2=\mathrm{GN}\big(\sigma(\mathbf{W}_2*\mathbf{x}_1+\mathbf{b}_2+e(\mathbf{u}))\big),
\end{equation}
with energy $E=\frac12\|\mathbf{x}_1-\hat{\mathbf{x}}_1\|^2+\frac12\|\mathbf{x}_2-\hat{\mathbf{x}}_2\|^2$. Here GN is a group norm with 8 groups and no affine parameters, whose variance is floored for CIFAR-100 and ImageNet32, and $\sigma(v)=\mathrm{SiLU}(\tau v)/\tau$. A sweep sets $\mathbf{x}_1\leftarrow\hat{\mathbf{x}}_1$ and then $\mathbf{x}_2\leftarrow\hat{\mathbf{x}}_2$, starting from zero. The decoder $d$ applies an affine group norm, global average pooling and a linear layer. Its input is $\mathbf{x}_2$ for the single readout and $\frac12(\mathbf{x}_1+\mathbf{x}_2)$ for the dual readout; because the decoder normalises its input first, this is the same as using the sum. After every optimiser step, the operator norm of each loop convolution on $8\times8$ maps is projected below a cap, using 10 power iterations. Encoder weights use He initialisation, and loop weights are drawn from $\mathcal{N}(0,0.1^2)$. For ImageNet32, the 7.24M parameters split into 2.01M in the encoder, 4.72M in the loop and 0.51M in the decoder.

\begin{table}[t]
\centering\small
\begin{tabular}{lccc}
\toprule
 & CIFAR-10/100 & ImageNet32 & output size \\
\midrule
encoder conv 1 & $3\times3$, 64, stride 1 & $5\times5$, 128, stride 1 & $32\times32$ \\
encoder conv 2 & $3\times3$, 128, stride 2 & $5\times5$, 256, stride 2 & $16\times16$ \\
encoder conv 3 & $3\times3$, 256, stride 2 & $3\times3$, 512, stride 2 & $8\times8$ \\
loop convs $\mathbf{W}_1, \mathbf{W}_2$ & $3\times3$, $256\to256$ & $3\times3$, $512\to512$ & $8\times8$ \\
\bottomrule
\end{tabular}
\caption{Convolutions of the circular model. Padding preserves the spatial size apart from the stride.}
\label{tab:vision-arch}
\end{table}

\paragraph{Learning rule.} The free phase runs Anderson acceleration (memory 9) on $\mathbf{x}_2$, followed by one sweep. Each training step then runs a single nudged phase with $\beta$ of random sign, warm-started from the free state. This phase is a coupled iteration of sweeps on the state and its prediction errors, in which the gradient of the cross-entropy enters at the readout node, or at both nodes for the dual readout. Since $\frac{\partial E}{\partial\theta}$ vanishes at the free equilibrium, the gradient estimate for the loop weights is $G_{\mathrm{EP}}(\beta)=\frac{1}{\beta}\frac{\partial E}{\partial\theta}(\theta,\mathbf{x}^{\beta})$, computed locally from the nudged prediction errors. As for Sudoku, a hinge penalty keeps the variance before the loop's group norm above a target.

\paragraph{Optimisation and evaluation.} We use Adam with gradient clipping at norm 1, L2 weight decay, a linear warm-up of 200 steps and cosine decay to zero. CIFAR-100 uses two cosine cycles of 90 epochs each. Table~\ref{tab:vision-hp} gives all hyper-parameters.  At test time, the state is relaxed from zero, with 40 Anderson steps for CIFAR or 40 sweeps for ImageNet32. No validation split is held out: we report the best test accuracy over epochs, averaged over seeds. Matrix products use TF32. Each run uses one H100 GPU, shared with at most one other job.

\begin{table}[t]
\centering\small
\begin{tabular}{lccc}
\toprule
 & CIFAR-10 & CIFAR-100 & ImageNet32 \\
\midrule
Width $w$ & 256 & 256 & 512 \\
Parameters & 1.55M & 1.58M & 7.24M \\
Batch size / epochs & 64 / 180 & 64 / 180 & 128 / 50 \\
Peak learning rate & $3.06\cdot10^{-4}$ & $7.24\cdot10^{-4}$ & $3.11\cdot10^{-4}$ \\
Weight decay (L2) & $6.74\cdot10^{-5}$ & $1.64\cdot10^{-4}$ & $2.6\cdot10^{-5}$ \\
Cosine cycles & 1 & 2 & 1 \\
$|\beta|$ & 2.21 & 3.43 & 3.17 \\
Free (Anderson) / nudged steps, training & 40 / 20 & 40 / 20 & 25 / 10 \\
$\tau$ & 1.632 & 1.340 & 1.627 \\
Spectral cap & 5.76 & 5.02 & 4.98 \\
GN variance floor & -- & 0.130 & 0.092 \\
Variance hinge (weight, target) & 2.89, 0.137 & 0.44, 0.130 & 0.035, 0.092 \\
Seeds & 4 & 4 & 4 per readout \\
\bottomrule
\end{tabular}
\caption{Hyper-parameters of the circular model.}
\label{tab:vision-hp}
\end{table}

\paragraph{Effective depth.} For Fig.~\ref{fig:cifar} (right), we relax each test image from zero for 60 plain sweeps, without Anderson acceleration, and read the prediction after every sweep. The switch sweep of an image is the first sweep from which its prediction stays correct up to sweep 60. The curves show the fraction of test images whose switch sweep is at most $t$. After $t$ sweeps, the input has passed through $3+2t$ layers.

\paragraph{VGG5 baseline.} Our VGG5 baseline follows \citet{pinchetti2025benchmarking}. It has four blocks of $3\times3$ convolution, $2\times2$ max-pooling and ReLU with 64, 128, 256 and 256 channels, followed by a linear layer, without normalisation. It has 963k parameters for CIFAR-10 and 987k for CIFAR-100. All layers are trained by EP with a random-sign nudge ($\beta=0.02$) on a squared-error cost. Training uses SGD with momentum $0.9$, learning rate $0.08$, weight decay $3\cdot10^{-4}$, cosine decay, batch size 64, 100 epochs and the same augmentation. We report the best test accuracy over training, averaged over three seeds.

\section{Sudoku: experimental details}
\label{app:sudoku}

\paragraph{Data.} We use the Sudoku-hard split of \citet{palm2018recurrent}: 180{,}000 training, 18{,}000 validation and 18{,}000 test puzzles, with the number of givens uniform between 17 and 34 (1{,}000 test puzzles for each value). During training, every sample is transformed by its own random symmetry of the puzzle: a permutation of the three row bands and of the rows inside each band, the same for column stacks and columns, a transposition with probability $1/2$, and a random relabelling of the nine digits. An \emph{epoch} is a random subset of $7{,}936$ training puzzles. Checkpoints are selected on the first $2{,}000$ validation puzzles. The test set is used only for the final evaluation.

\paragraph{Chain depth.} The reference solver never sees the network. At each round, it finds every blank cell that is forced either by a \emph{naked single} (only one digit is compatible with the cell's row, column and box) or by a \emph{hidden single} (the cell is the only place left for some digit in a row, column or box), and fills all of them at once. If no cell is forced, it writes the true digit into a cell with the fewest candidates, and counts this as a round. The chain depth of a board is the number of rounds needed to complete it. The two rules alone complete $88.5\%$ of the test boards, and the median depth is $7$. Every test board has depth at least $3$. The five groups of Fig.~\ref{fig:sudoku} contain 2{,}431 (3--4), 4{,}871 (5--6), 4{,}406 (7--9), 3{,}829 (10--14) and 2{,}463 ($\geq 15$) boards.

\paragraph{Circular block.} The input of a cell is a one-hot code of its content (blank or one of nine digits) concatenated with one-hot codes of its row, column and box, 37 features in total, which we stack into $\mathbf{u}\in\mathbb{R}^{81\times37}$. A linear layer maps them to $e(\mathbf{u})=\mathbf{u}\mathbf{W}_e+\mathbf{b}_e\in\mathbb{R}^{81\times w}$. The loop has four value nodes, $\mathbf{x}_1\in\mathbb{R}^{w\times 162}$, $\mathbf{x}_2\in\mathbb{R}^{81\times w}$, $\mathbf{x}_3\in\mathbb{R}^{81\times 2w}$ and $\mathbf{x}_4\in\mathbb{R}^{81\times w}$, with predictions
\begin{equation}
\begin{aligned}
\hat{\mathbf{x}}_1 &= \sigma\big(\mathrm{LN}(\mathbf{x}_4)^{\top}\mathbf{W}_1+\mathbf{b}_1\big), &
\hat{\mathbf{x}}_2 &= \big(\mathbf{x}_1\mathbf{W}_2+\mathbf{b}_2\big)^{\top},\\
\hat{\mathbf{x}}_3 &= \sigma\big(\mathrm{LN}(\mathbf{x}_2)\mathbf{W}_3+\mathbf{b}_3\big), &
\hat{\mathbf{x}}_4 &= \mathbf{x}_3\mathbf{W}_4+\mathbf{b}_4+e(\mathbf{u}),
\end{aligned}
\end{equation}
and energy $E=\tfrac12\sum_{i=1}^{4}\|\mathbf{x}_i-\hat{\mathbf{x}}_i\|^2$. The first pair mixes across cells, and the second mixes across channels within each cell. $\mathrm{LN}$ is a layer norm over channels without affine parameters (variance floor $2.06\cdot10^{-3}$), $\sigma(v)=\mathrm{GELU}(\tau v)/\tau$ with $\tau=2.358$, and there is no skip connection. A sweep updates $\mathbf{x}_1,\mathbf{x}_2,\mathbf{x}_3,\mathbf{x}_4$ in this order, setting each node to its prediction, starting from zero. The decoder $d$ applies an affine layer norm and a linear map $\mathbb{R}^{w}\to\mathbb{R}^{9}$ to every cell of $\mathbf{x}_4$. The output cost $C$ is the cross-entropy averaged over the 81 cells. The mixing weights are initialised from $\mathcal{N}\big(0,(0.25/\sqrt{\mathrm{fan_{in}}})^2\big)$ and the biases at zero. After every optimiser step, the spectral norms of $\mathbf{W}_1,\dots,\mathbf{W}_4$ are projected below a cap, using 8 power iterations.

\paragraph{Learning rule.} All weights of the loop, including the embedding $e$, are trained with EP; nothing is backpropagated. From the free equilibrium $\mathbf{x}^*$, two nudged phases minimise $E\pm\beta C$. The gradient estimate is the centred difference $G_{\mathrm{EP}}^{\mathrm{c}}(\beta)=\frac{1}{2\beta}\big[\frac{\partial E}{\partial\theta}(\theta,\mathbf{x}^{\beta})-\frac{\partial E}{\partial\theta}(\theta,\mathbf{x}^{-\beta})\big]$ of Eq.~\eqref{eq:ep_centered}. The embedding $e$ receives its gradient through $e(\mathbf{u})$, which enters $E$ only through $\hat{\mathbf{x}}_4$. The decoder, a single normalised linear layer, is trained by the gradient of $C$ at the free equilibrium. A hinge penalty keeps the variance of the predictions above a fixed target. All runs use Adam (moment decay rates $0.9$ and $0.999$) with gradient clipping at norm $1$ and L2 weight decay. The Adam updates of the token- and channel-mixing weights are scaled by $0.918$ and $0.435$. Table~\ref{tab:sudoku-hp} lists the remaining settings.

\paragraph{Standard model.} During training, the free phase runs 150 steps of Anderson acceleration (memory 13) on $\mathbf{x}_4$, followed by one sweep. The nudged phases start from the free state and iterate the state and its prediction errors jointly for 5 steps, with Anderson acceleration (memory 2). Samples whose free phase has not converged (relative residual $\geq10^{-3}$) are left out of the update. The spectral cap is adapted once per epoch, within $[0.30, 0.70]$, by a proportional controller that keeps the spectral radius of the loop's Jacobian $\mathbf{J}_f$ near $1.25$. The radius is measured with 80 sweeps on 256 validation puzzles. At test time, we run sweeps from zero with the readout node damped, $\mathbf{x}_4\leftarrow\frac12\mathbf{x}_4+\frac12\hat{\mathbf{x}}_4$, which removes a period-two oscillation of the undamped iteration. The prediction is read after every sweep. Checkpoints are selected at 300 sweeps; Fig.~\ref{fig:sudoku} continues the same iteration to 1500 sweeps.

\paragraph{Recursive model.} Stage $s$ uses the input
$e(\mathbf{u})+\mathrm{LN}\big(\mathbf{a}^{(s-1)}\big)\mathbf{W}_a$,
where $\mathbf{a}^{(s)}$ is the free equilibrium of $\mathbf{x}_4$ at stage $s$ and $\mathbf{a}^{(0)}=\mathbf{0}$. The feedback matrix $\mathbf{W}_a\in\mathbb{R}^{w\times w}$ is initialised at zero and holds $14{,}400$ of the $104{,}736$ parameters. Each stage is warm-started from the whole free state of the previous one. The free phase is Anderson acceleration (memory 13) for 600 steps at the first stage and 80 at later ones, in training and at test time alike. The nudged phases use 500 heavy-ball steps (step size $0.0208$, momentum $0.992$). Each batch is unrolled for $S\sim\mathcal{U}\{4,\dots,16\}$ stages. Every stage computes its own EP update and takes its own optimiser step, and $\mathbf{a}$ is a constant in every update. From the fifth stage on, the nudged phases run on 64 of the 128 samples. A quarter of these slots go to samples already solved at the previous stage and the rest to unsolved ones; this uses the labels and is done only in training. We keep an exponential moving average of the weights (decay $0.999$), which is the model we evaluate. At test time, we run 32 stages and read the prediction after each.

\paragraph{Relation to TRMs.} A TRM \citep{jolicoeur2025less} keeps an answer $\mathbf{a}$ and a latent $\mathbf{h}$. Each cycle updates $\mathbf{h}$ several times with a small network and then $\mathbf{a}$ once. The loss is backpropagated through the last cycle, and $(\mathbf{a},\mathbf{h})$ are detached between supervision steps. In our model, the relaxation of the loop replaces the latent updates, the readout node at equilibrium is the answer, and $\mathbf{W}_a$ is the only path that carries the answer to the next stage. We keep from TRMs:
\begin{itemize}
  \item deep supervision with a detached carry and an optimiser step after every supervision step;
  \item a moving average of the weights;
  \item symmetry augmentation;
  \item an attention-free mixer.
\end{itemize}
We drop three of its ingredients:
\begin{itemize}
  \item backpropagation, since every stage is a local EP contrast;
  \item the cycles without gradient: supervising only the last stage made the chain collapse after five stages;
  \item the learned halting, since the number of stages is random in training and fixed at test time.
\end{itemize}

\begin{table}[t]
\centering\small
\begin{tabular}{lcc}
\toprule
 & Standard & Recursive \\
\midrule
Width $w$ (token / channel hidden) & 512 (162 / 1024) & 120 (162 / 240) \\
Parameters & 1{,}101{,}696 & 104{,}736 \\
Batch size & 256 & 128 \\
Epochs ($7{,}936$ puzzles each) & 250 & 90 \\
Optimiser steps & 7{,}750 & $\approx$55{,}800 (one per stage) \\
Peak learning rate, warm-up & $5.2\cdot10^{-4}$, 186 steps & $3\cdot10^{-4}$, 100 steps \\
Schedule & cosine to 0 in 60 epochs, then $6\cdot10^{-5}$ & cosine to $10^{-6}$ \\
Weight decay (L2) & $1.21\cdot10^{-4}$ & $1.21\cdot10^{-4}$ \\
$\beta$ & 0.166 & 0.02 \\
Free phase (training) & Anderson, 150 steps & Anderson, 600 / 80 steps \\
Nudged phase & coupled Anderson, 5 steps & heavy-ball, 500 steps \\
Spectral cap & adaptive, $[0.30, 0.70]$ & 0.7 \\
Variance hinge (weight, target) & 2.92, 0.099 & 2.92, 0.144 \\
Weight average & -- & EMA 0.999 \\
Matrix products & TF32 & FP32 \\
Checkpoint selection & 300 sweeps & 16 stages \\
Seeds (selected epoch) & 301, 302, 303 (106, 118, 157) & 275, 293, 294 (82, 86, 86) \\
Training time, one H100 GPU & 0.95 h & 7.3 h \\
\bottomrule
\end{tabular}
\caption{Sudoku hyper-parameters.}
\label{tab:sudoku-hp}
\end{table}

\paragraph{Evaluation.} A board is correct if all 81 predicted digits match the solution. For the multi-view results, we fix 1024 digit permutations, the identity and the others distinct random ones, and use the same permutations for every board and both models. Each view is solved independently, and its prediction is mapped back to the original digits. We select the view with the lowest self-cost $C_{\mathrm{self}}=-\frac{1}{81}\sum_{p}\max_c\log\mathrm{softmax}(\mathbf{x}_{\mathcal{O},p})_c$, where $\mathbf{x}_{\mathcal{O},p}=d(\mathbf{x}_4)_p\in\mathbb{R}^9$ are the output logits of cell $p$. This is the output cost of the model's own answer, i.e.\ the cost that the nudge would minimise if the prediction were the target. Ties go to the first view, and the selection is repeated at every sweep or stage. Single-view numbers use all 18{,}000 test boards. The 1024-view numbers and panels (c, g) use a stratified subset of 5{,}000 test boards, proportional to the five groups and drawn with a fixed seed. The solve iteration of a board is the first sweep (or stage) from which it stays correct until the last one. Unsolved boards are excluded. $\rho$ is the Spearman correlation between solve iteration and chain depth, computed for each seed and averaged over seeds. 

\paragraph{Inference cost.} Counting the multiply-adds of one sweep from the layer shapes gives about 98M per board for the standard loop and 7.8M for the recursive one. Per view, the standard model at 500 sweeps uses about 49G multiply-adds. The recursive model at 32 stages runs $600+31\times80$ free-phase steps and uses about 24G. Both counts ignore the small overhead of Anderson acceleration.
\newpage

\section{Language modeling: experimental details}
\label{app:lm}

\textbf{Data.} \emph{TinyStories.} We use TinyStoriesV2-GPT4 \citep{eldan2023tinystories}. A byte-level BPE with a vocabulary of $8192$ tokens is trained once on a $100$ MB sample of the training split. The training set is the first $200$ MB of the training split, i.e.\ $48.1$M tokens. \emph{WikiText-103.} We use the raw version of WikiText-103 \citep{merity2017wiki103}, in which every article is followed by an end-of-text token. A byte-level BPE with a vocabulary of $8192$ tokens is trained on a $200$ MB sample of the training split, which then contains $134.7$M tokens. To match the size of the TinyStories training set, we train on its first $48.1$M tokens, i.e.\ on the first $36\%$ of the articles; the official validation and test splits are kept whole. In both datasets, at every step we draw $32$ windows of $129$ tokens at random offsets of the training token stream, which give $32$ sequences of $128$ inputs and next-token targets. Validation perplexity is measured on a fixed set of $50 \times 16$ windows of $128$ tokens ($102{,}400$ tokens) of the validation split, drawn once with a fixed seed and shared by all runs. For WikiText-103, test perplexity is measured on the entire official test split ($0.32$M tokens), cut into $2515$ non-overlapping windows of $128$ tokens, each relaxed independently from the zero state. Perplexity is the exponential of the cross-entropy averaged over all target tokens. Since the vocabulary is a subword one, the WikiText-103 perplexities are per token and are not comparable with the word-level values usually reported on this benchmark.

\textbf{Architecture.} Both datasets use the same loop: a single causal transformer block with $L = 3$ value nodes $x_1, x_3 \in \mathbb{R}^{T \times w}$ and $x_2 \in \mathbb{R}^{T \times 4w}$, and energy $E = \frac{1}{2}\sum_{i=1}^{3} \lVert x_i - \hat{x}_i \rVert^2$. A sweep sets $x_1 \leftarrow \hat{x}_1$, $x_2 \leftarrow \hat{x}_2$, and then $x_3 \leftarrow \hat{x}_3$, starting from zero, and the decoder $d$ reads $x_3$. The input $e(u)$ is the sum of a token and a position embedding, normalised per token and rescaled to $0.5$. The predictions are
\begin{equation}
  \hat{x}_1 = \mathrm{Attn}\big(\mathrm{LN}(x_3), e(u)\big)\, W_o + e(u)\, W_s, \qquad
  \hat{x}_2 = \mathrm{LN}(x_1)\, W_1 + b_1, \qquad
  \hat{x}_3 = \mathrm{GELU}(x_2)\, W_2 + b_2,
  \label{eq:lm-ts}
\end{equation}
where LN is a layer norm without affine parameters whose variance is floored at $0.1$, so that its gain is at most $1/\sqrt{0.1}$. The queries, keys and values of the causal multi-head attention are read from $\mathrm{LN}(x_3)$, to which a linear map of $e(u)$ is added. Queries and keys are softly normalised per head, $q \leftarrow q / (\lVert q \rVert^2 + d_h)^{1/2}$, and the logits are $4\, q^\top k$, so that they lie in $[-4, 4]$. The skip term $e(u)\, W_s$ gives every position direct access to its own token; since it does not depend on the state, it does not change the Jacobian of the loop. The width is $w = 384$ with $6$ heads, the context is $128$ tokens, and there is no residual connection. After every optimiser step, the spectral norms of $W_q, W_k, W_v, W_o, W_1, W_2$ are projected below $1$, using $10$ power iterations. Weights are initialised from $\mathcal{N}(0, 1/\text{fan-in})$ and biases at zero. The decoder is an affine layer norm followed by a linear map to the vocabulary, and the input and output embeddings are not tied.

\textbf{Learning rule.} All weights of the loop and the input embeddings are trained with EP; the embeddings receive their gradient through $e(u)$, which enters $E$ only through $\hat{x}_1$. The free phase runs Anderson acceleration (memory $6$) from zero for $20$ iterations. The nudged phase starts from the free equilibrium $x^*$ and takes the Newton step on $F_\beta = E + \beta C$ described in Section~\ref{sec:lm}, using the curvature of $E$ at $x^*$. The step is computed with $150$ (TinyStories) or $400$ (WikiText-103) iterations of the heavy-ball method, which only uses Hessian--vector products. Its step size and momentum are set by a lower bound on the curvature, $0.1$, and an upper bound, estimated at every training step from ten Hessian--vector products, as the maximum over the batch with a margin of $1.3$. On WikiText-103, $150$ iterations are not enough: after a few thousand steps the upper bound on the curvature grows, the step is no longer computed to convergence, and training stalls; with $400$ iterations the curvature stays bounded and the step converges throughout training. The gradient estimate is the centred difference $G^{c}_{\mathrm{EP}}(\beta)$ of Eq.~(4) with $\beta = 0.1$, and the cost $C$ is the cross-entropy averaged over the target tokens of the batch. Samples whose nudged response exceeds ten times the batch median are left out of the update. A hinge penalty on the spectral radius of the loop Jacobian at the free equilibrium keeps it below $0.95$. The decoder is trained by the gradient of $C$ at the free equilibrium (Section~\ref{sec:lm}).

\textbf{DEQ baseline.} The DEQ has the same predictions, initialisation, data and optimiser as the CPCN, and is trained with the implicit gradient of its fixed point. The forward pass runs $30$ plain sweeps from zero, and the backward pass solves the adjoint equation $u = J^\top u + g$ at the fixed point with $25$ plain iterations, followed by one vector--Jacobian product with respect to the parameters.

\textbf{Optimisation and evaluation.} We use AdamW with decoupled weight decay $0.01$ (not applied to embeddings, biases and layer-norm gains), gradient clipping at global norm $0.25$, a linear warm-up of $1000$ steps to a learning rate of $2.5 \cdot 10^{-4}$, and cosine decay to zero at $20{,}000$ steps. Table~\ref{tab:lm-hparams} lists all settings. At test time, the state is relaxed from zero with Anderson acceleration, for $20$ iterations for CPCNs and $30$ for the DEQ, which suffices for convergence of every model. We report the final checkpoint, without selection, averaged over three seeds that set the initialisation and the order of the data. Each run uses one H100 GPU.

\textbf{Effective depth.} For the right panel of Fig.~\ref{fig:lm_depth}, we relax each validation sequence from zero for $100$ plain sweeps, without Anderson acceleration, and record the loss of every token after every sweep. Its converged loss is computed with $200$ Anderson iterations. A token is committed after sweep $t$ if its loss stays within $0.01$ nats of the converged value from sweep $t$ up to sweep $100$. After $t$ sweeps, the input has passed through $t$ transformer blocks. For the left panel, the spectral radius of the Jacobian of one sweep with respect to $x_3$ is estimated at the free equilibrium by $60$ power iterations, and averaged over $200$ validation sequences, at initialisation and every $1000$ training steps.

\begin{table}[H]
\centering
\begin{tabular}{lcc}
\toprule
 & TinyStories & WikiText-103 \\
\midrule
Width $w$ / heads / MLP hidden & 384 / 6 / 1536 & 384 / 6 / 1536 \\
Context, vocabulary & 128, 8192 & 128, 8192 \\
Parameters & 8.71M & 8.71M \\
Training tokens & 48.1M & 48.1M (of 134.7M) \\
Batch size / steps & 32 / 20,000 & 32 / 20,000 \\
Peak learning rate, warm-up & $2.5 \cdot 10^{-4}$, 1000 steps & $2.5 \cdot 10^{-4}$, 1000 steps \\
Weight decay (decoupled), gradient clip & 0.01, 0.25 & 0.01, 0.25 \\
$\beta$ & 0.1 & 0.1 \\
Free phase (Anderson), training and test & 20 & 20 \\
Nudged phase (heavy-ball iterations) & 150 & 400 \\
Curvature lower bound & 0.1 & 0.1 \\
Spectral cap & 1 & 1 \\
Layer-norm variance floor & 0.1 & 0.1 \\
Attention logit bound & 4 & 4 \\
Spectral-radius hinge (weight, target) & 1, 0.95 & 1, 0.95 \\
Matrix products & FP32 & FP32 \\
DEQ: forward / adjoint iterations & 30 / 25 (plain) & 30 / 25 (plain) \\
Reported perplexity & validation & test \\
Seeds & 0, 1, 2 & 0, 1, 2 \\
Time per run, one H100 GPU (CPCN / DEQ) & 1.8 h / 0.25 h & 3.8 h / 0.25 h \\
\bottomrule
\end{tabular}
\caption{Hyper-parameters of the language models.}
\label{tab:lm-hparams}
\end{table}

\section{Formal statements and proofs}
\label{app:theory}

This appendix contains the precise versions of the two statements of
the main body, together with their proofs. 
\subsection{Setting}
\label{app:setting}

\paragraph{The graph and its state.}
The vertex set is $\mathcal{V}=\{\mathcal{I},1,\ldots,L,\mathcal{O}\}$. The vertices
$1,\ldots,L$ form the loop and carry value nodes
$\mathbf{x}_i\in\mathbb{R}^{d_i}$; the input vertex is clamped to the observation
$\mathbf{u}$ and is therefore not a variable, and the output vertex carries
$\mathbf{x}_{\mathcal{O}}\in\mathbb{R}^{d_{\mathcal{O}}}$. To lighten
Eq.~\ref{eq:cpc-pred} we collect the edge maps of the loop into
\begin{equation}
    \phi_1 \coloneqq g_{L,1},
    \qquad
    \phi_i \coloneqq g_{i-1,i}
    \quad (2\leq i\leq L-1),
    \qquad
    \phi_L \coloneqq g_{L-1,L}\bigl(\,\cdot\,,e(\mathbf{u})\bigr),
    \label{eq:phi}
\end{equation}
so that $\phi_i:\mathbb{R}^{d_{i-1}}\to\mathbb{R}^{d_i}$ for every $i$, under the
cyclic convention $d_0\coloneqq d_L$; the dependence of $\phi_L$ on $\theta$ and
$\mathbf{u}$ is left implicit. We write
\begin{equation}
    \mathbf{x}_{\mathrm{loop}}
    =(\mathbf{x}_1,\ldots,\mathbf{x}_L)\in\mathbb{R}^{n},
    \quad
    n=\sum_{i=1}^{L}d_i,
    \qquad
    \mathbf{x}
    =(\mathbf{x}_{\mathrm{loop}},\mathbf{x}_{\mathcal{O}})
    \in\mathbb{R}^{n+d_{\mathcal{O}}}
\end{equation}
for the state of the loop and for the vector of all free value nodes.

\paragraph{The energy.}
With the predictions of Eq.~\ref{eq:cpc-pred}, the energy of Eq.~\ref{eq:energy}
splits into a loop part and a single decoder term,
\begin{equation}
    E(\theta,\mathbf{x})
    =
    \underbrace{
        \frac{1}{2}\sum_{i=1}^{L}
        \bigl\|\mathbf{x}_i-\hat{\mathbf{x}}_i\bigr\|_2^2
    }_{
        \eqqcolon\,E_{\mathrm{loop}}(\theta,\mathbf{x}_{\mathrm{loop}})
    }
    \;+\;
    \frac{1}{2}\bigl\|
        \mathbf{x}_{\mathcal{O}}-d(\mathbf{x}_L)
    \bigr\|_2^2 ,
    \label{eq:app-split}
\end{equation}
the first group involving $\mathbf{x}_{\mathcal{O}}$ through no term at
all.\footnote{The input vertex has no parents, so its prediction is an empty sum and
its error term is $\tfrac12\|\mathbf{u}\|_2^2$, a constant independent of both the
free state and the parameters. We discard it throughout, which affects no stationarity
condition and no parameter gradient.}

\paragraph{The two maps.}
Because all value nodes are updated at the same time, the object attached to the loop
by the inference dynamics is the \emph{simultaneous prediction map}
\begin{equation}
    S_\theta(\cdot,\mathbf{u}):\mathbb{R}^{n}\to\mathbb{R}^{n},
    \qquad
    S_\theta(\mathbf{x}_{\mathrm{loop}},\mathbf{u})
    =\bigl(
        \phi_1(\mathbf{x}_L),\,
        \phi_2(\mathbf{x}_1),\,
        \ldots,\,
        \phi_L(\mathbf{x}_{L-1})
     \bigr),
    \label{eq:S}
\end{equation}
which stacks every local prediction evaluated at the current state, whereas the deep
equilibrium model of Section~\ref{sec:deq} iterates the composition around the loop,
\begin{equation}
    f_\theta(\cdot,\mathbf{u}):\mathbb{R}^{d_L}\to\mathbb{R}^{d_L},
    \qquad
    f_\theta(\cdot,\mathbf{u})
    =\phi_L\circ\phi_{L-1}\circ\cdots\circ\phi_1 .
    \label{eq:T}
\end{equation}
The first acts on the joint state, the second on the main value node alone;
Lemma~\ref{lem:fp} relates them. Since the $i$-th block of
$\mathbf{x}_{\mathrm{loop}}-S_\theta(\mathbf{x}_{\mathrm{loop}},\mathbf{u})$ is the
prediction error $\boldsymbol{\varepsilon}_i$, the loop energy is exactly the squared
residual of the simultaneous map,
\begin{equation}
    E_{\mathrm{loop}}(\theta,\mathbf{x}_{\mathrm{loop}})
    =\frac{1}{2}\bigl\|
        \mathbf{r}(\mathbf{x}_{\mathrm{loop}})
     \bigr\|_2^2,
    \qquad
    \mathbf{r}(\mathbf{x}_{\mathrm{loop}})
    \coloneqq
    \mathbf{x}_{\mathrm{loop}}
    -S_\theta(\mathbf{x}_{\mathrm{loop}},\mathbf{u}) .
    \label{eq:app-residual}
\end{equation}
This identity is what the whole of Section~\ref{app:inference} rests on.

\begin{remark}[The parameter gradient carries the encoder]
\label{rem:encoder}
Throughout, $\nabla_\theta E(\theta,\mathbf{x})$ denotes the gradient of the map
$\theta\mapsto E(\theta,\mathbf{x})$ at fixed state and fixed observation
$\mathbf{u}$. By Eq.~\ref{eq:phi} this includes the path through the injection
$e(\mathbf{u})$, which enters $\hat{\mathbf{x}}_L$ and hence
$\boldsymbol{\varepsilon}_L$. Treating that injection as a precomputed constant would
differentiate a different function, one in which the encoder parameters do not appear
at all; Theorem~\ref{thm:ep-app} then applies to that other function and says nothing
about $e$.
\end{remark}

\paragraph{Assumption}\label{ass:A1}
Every map of Eq.~\ref{eq:phi} is continuously differentiable and $q$-Lipschitz with
respect to the Euclidean norm, for a common modulus $q<1$, the map $\phi_L$ being
required to be so in its first argument at fixed $\mathbf{u}$. The decoder $d$
is continuously differentiable, and no contractivity is asked of it.

Three lemmas precede the results. The first eliminates the vertices outside the loop,
the second turns Assumption~\ref{ass:A1} into a bound on the joint map, and the third
identifies the fixed points of the two maps of Eqs.~\ref{eq:S} and \ref{eq:T}.

\begin{lemma}[Elimination of the output vertex]
\label{lem:out}
A state $\mathbf{x}=(\mathbf{x}_{\mathrm{loop}},\mathbf{x}_{\mathcal{O}})$ is a
stationary point of $E(\theta,\cdot)$ in the free value nodes if and only if
\begin{equation}
    \nabla_{\mathbf{x}_{\mathrm{loop}}}
    E_{\mathrm{loop}}(\theta,\mathbf{x}_{\mathrm{loop}})=0
    \qquad\text{and}\qquad
    \mathbf{x}_{\mathcal{O}}=d(\mathbf{x}_L),
\end{equation}
and at any such state
$E(\theta,\mathbf{x})=E_{\mathrm{loop}}(\theta,\mathbf{x}_{\mathrm{loop}})$.
\end{lemma}

\begin{proof}
By Eq.~\ref{eq:app-split}, stationarity in $\mathbf{x}_{\mathcal{O}}$ reads
$\boldsymbol{\varepsilon}_{\mathcal{O}}
=\mathbf{x}_{\mathcal{O}}-d(\mathbf{x}_L)=0$. The output vertex has no
children, so it enters no prediction, and the only place it touches the remaining
stationarity conditions is the term
$-\mathbf{J}_{d}(\mathbf{x}_L)^{\!\top}\boldsymbol{\varepsilon}_{\mathcal{O}}$
of $\partial E/\partial\mathbf{x}_L$, which vanishes together with
$\boldsymbol{\varepsilon}_{\mathcal{O}}$. The conditions on the loop variables reduce
accordingly to those of $E_{\mathrm{loop}}$. The converse substitution is immediate,
as is the value of the energy, the decoder term being zero.
\end{proof}

\begin{lemma}[Contraction of the simultaneous map]
\label{lem:contraction}
Under Assumption~\ref{ass:A1}, the map $S_\theta(\cdot,\mathbf{u})$ is $q$-Lipschitz on
$\mathbb{R}^n$, and
$\|\mathbf{J}_S(\mathbf{x}_{\mathrm{loop}})\|_2\leq q$ at every state.
\end{lemma}

\begin{proof}
Let $P$ be the coordinate permutation of $\mathbb{R}^n$ sending
$(\mathbf{x}_1,\ldots,\mathbf{x}_L)$ to
$(\mathbf{x}_L,\mathbf{x}_1,\ldots,\mathbf{x}_{L-1})$, so that the $i$-th block of
$P\mathbf{x}_{\mathrm{loop}}$ is $\mathbf{x}_{i-1}$ and lies in $\mathbb{R}^{d_{i-1}}$,
the domain of $\phi_i$. Letting $\boldsymbol{\phi}$ act blockwise on that
decomposition,
$\boldsymbol{\phi}(\mathbf{w})=(\phi_1(\mathbf{w}_1),\ldots,\phi_L(\mathbf{w}_L))$, a
comparison with Eq.~\ref{eq:S} gives the factorization
\begin{equation}
    S_\theta(\cdot,\mathbf{u})=\boldsymbol{\phi}\circ P .
    \label{eq:factorization}
\end{equation}
For any $\mathbf{w},\mathbf{w}'$,
\begin{equation}
    \bigl\|\boldsymbol{\phi}(\mathbf{w})-\boldsymbol{\phi}(\mathbf{w}')\bigr\|_2^2
    =\sum_{i=1}^{L}\bigl\|\phi_i(\mathbf{w}_i)-\phi_i(\mathbf{w}'_i)\bigr\|_2^2
    \leq q^2\sum_{i=1}^{L}\|\mathbf{w}_i-\mathbf{w}'_i\|_2^2
    =q^2\|\mathbf{w}-\mathbf{w}'\|_2^2 ,
\end{equation}
so $\boldsymbol{\phi}$ is $q$-Lipschitz, and $P$ is orthogonal; the first claim
follows from Eq.~\ref{eq:factorization}. Differentiating that same factorization gives
$\mathbf{J}_S=\mathbf{J}_{\boldsymbol{\phi}}P$ with
$\mathbf{J}_{\boldsymbol{\phi}}
=\operatorname{blockdiag}(\mathbf{J}_{\phi_1},\ldots,\mathbf{J}_{\phi_L})$, whence
\begin{equation}
    \|\mathbf{J}_S\|_2
    \leq\|\mathbf{J}_{\boldsymbol{\phi}}\|_2\,\|P\|_2
    =\max_{1\leq i\leq L}\|\mathbf{J}_{\phi_i}\|_2
    \leq q ,
\end{equation}
the last bound holding pointwise because a differentiable $q$-Lipschitz map has
Jacobian of operator norm at most $q$.
\end{proof}

\begin{lemma}[The two maps have the same fixed points]
\label{lem:fp}
The assignment $\mathbf{x}_{\mathrm{loop}}\mapsto\mathbf{x}_L$ is a bijection from the
fixed points of $S_\theta(\cdot,\mathbf{u})$ onto those of $f_\theta(\cdot,\mathbf{u})$, with
inverse
$\mathbf{z}\mapsto
\bigl(\phi_1(\mathbf{z}),\,\phi_2(\phi_1(\mathbf{z})),\,\ldots,\,\mathbf{z}\bigr)$.
\end{lemma}

\begin{proof}
If $S_\theta(\mathbf{x}_{\mathrm{loop}},\mathbf{u})=\mathbf{x}_{\mathrm{loop}}$ then
$\mathbf{x}_1=\phi_1(\mathbf{x}_L)$ and $\mathbf{x}_i=\phi_i(\mathbf{x}_{i-1})$ for
$2\leq i\leq L$; substituting the first $L-1$ identities into the last one gives
$\mathbf{x}_L=(\phi_L\circ\cdots\circ\phi_1)(\mathbf{x}_L)
=f_\theta(\mathbf{x}_L,\mathbf{u})$. Conversely, if
$f_\theta(\mathbf{z},\mathbf{u})=\mathbf{z}$, the state defined by
$\mathbf{x}_1=\phi_1(\mathbf{z})$, by $\mathbf{x}_i=\phi_i(\mathbf{x}_{i-1})$ for
$2\leq i\leq L-1$, and by $\mathbf{x}_L=\mathbf{z}$, satisfies every block of the
fixed-point condition of $S_\theta(\cdot,\mathbf{u})$, the last one precisely because
$\mathbf{z}$ is fixed by the composition. The two constructions are mutually inverse.
\end{proof}

\subsection{Inference}
\label{app:inference}

\begin{theorem}[Formal version of Theorem 1]
\label{thm:unique-app}
Let Assumption~\ref{ass:A1} hold. Then $E(\theta,\cdot)$ has exactly one stationary
point $\mathbf{x}^*$ in the free value nodes, and
\begin{enumerate}[label=(\roman*),leftmargin=*,topsep=2pt,itemsep=1pt]
    \item $E(\theta,\mathbf{x}^*)=0$, and $\mathbf{x}^*$ is the unique global
          minimizer of $E(\theta,\cdot)$;
    \item $\mathbf{x}^*_L$ is the unique fixed point of $f_\theta(\cdot,\mathbf{u})$,
          and the remaining value nodes are obtained from it by a single pass around
          the loop, namely $\mathbf{x}^*_1=\phi_1(\mathbf{x}^*_L)$ and
          $\mathbf{x}^*_i=\phi_i(\mathbf{x}^*_{i-1})$ for $2\leq i\leq L-1$,
          together with $\mathbf{x}^*_{\mathcal{O}}=d(\mathbf{x}^*_L)$;
    \item every trajectory of the gradient flow
          $\dot{\mathbf{x}}=-\nabla_{\mathbf{x}}E(\theta,\mathbf{x})$ that converges
          to a stationary point converges to $\mathbf{x}^*$.
\end{enumerate}
\end{theorem}

\begin{proof}
By Lemma~\ref{lem:out} it suffices to prove the corresponding statements for
$E_{\mathrm{loop}}$, the output value node being slaved to $\mathbf{x}_L$.

\emph{Existence, and vanishing energy.}
By Lemma~\ref{lem:contraction} the map $S_\theta(\cdot,\mathbf{u})$ is a contraction of the
complete space $\mathbb{R}^n$, so the Banach fixed-point theorem provides a unique
$\mathbf{x}^*_{\mathrm{loop}}$ with
$S_\theta(\mathbf{x}^*_{\mathrm{loop}},\mathbf{u})
=\mathbf{x}^*_{\mathrm{loop}}$. There $\mathbf{r}=0$, hence
$E_{\mathrm{loop}}=0$ by Eq.~\ref{eq:app-residual}, and since the energy is
nonnegative everywhere this state is a global minimizer; it is the only one, a squared
norm vanishing exactly where its argument does.

\emph{Uniqueness of the stationary point.}
Differentiating Eq.~\ref{eq:app-residual},
\begin{equation}
    \nabla_{\mathbf{x}_{\mathrm{loop}}}
    E_{\mathrm{loop}}(\theta,\mathbf{x}_{\mathrm{loop}})
    =\bigl(\mathbf{I}-\mathbf{J}_S(\mathbf{x}_{\mathrm{loop}})\bigr)^{\!\top}
     \mathbf{r}(\mathbf{x}_{\mathrm{loop}}) .
    \label{eq:app-grad}
\end{equation}
By Lemma~\ref{lem:contraction} we have $\|\mathbf{J}_S\|_2\leq q<1$ at every state, so
$\mathbf{I}-\mathbf{J}_S$ is invertible everywhere, its inverse being the convergent
Neumann series $\sum_{k\geq0}\mathbf{J}_S^{\,k}$. Hence Eq.~\ref{eq:app-grad} vanishes
if and only if $\mathbf{r}(\mathbf{x}_{\mathrm{loop}})=0$, that is, exactly at the
fixed point just constructed. This proves both the uniqueness and item (i).

\emph{Identification with the deep equilibrium model.}
Item (ii) is Lemma~\ref{lem:fp} applied to $\mathbf{x}^*_{\mathrm{loop}}$, the
fixed point of $f_\theta(\cdot,\mathbf{u})$ being unique because that of
$S_\theta(\cdot,\mathbf{u})$ is and the correspondence is bijective, together with
Lemma~\ref{lem:out} for the output vertex.

\emph{Gradient flow.}
A convergent trajectory has a stationary limit, and there is only one, which is item
(iii).
\end{proof}

The next two statements sharpen the theorem in the two directions one would want, and
both come essentially for free from the residual form of the energy: the first
strengthens the convergence of the free phase from conditional to global, the second
shows that the equilibrium so reached is nondegenerate, which is precisely the
hypothesis under which the learning result is proved.

\begin{proposition}[Quadratic growth and global convergence]
\label{prop:coercive}
Under Assumption~\ref{ass:A1},
\begin{equation}
    E_{\mathrm{loop}}(\theta,\mathbf{x}_{\mathrm{loop}})
    \;\geq\;
    \frac{(1-q)^2}{2}
    \bigl\|\mathbf{x}_{\mathrm{loop}}-\mathbf{x}^*_{\mathrm{loop}}\bigr\|_2^2
    \qquad
    \text{for every }\mathbf{x}_{\mathrm{loop}}\in\mathbb{R}^n ,
    \label{eq:app-growth}
\end{equation}
and $E(\theta,\cdot)$ has bounded sublevel sets. If moreover the maps of
Eq.~\ref{eq:phi} and the decoder are twice continuously differentiable, so that
$\nabla_{\mathbf{x}}E$ is locally Lipschitz, then the gradient flow is defined for all
$t\geq0$ from any initial state and converges to $\mathbf{x}^*$.
\end{proposition}

\begin{proof}
Using $\mathbf{x}^*_{\mathrm{loop}}
=S_\theta(\mathbf{x}^*_{\mathrm{loop}},\mathbf{u})$ and
Lemma~\ref{lem:contraction},
\begin{align}
    \bigl\|\mathbf{r}(\mathbf{x}_{\mathrm{loop}})\bigr\|_2
    &=\Bigl\|
        (\mathbf{x}_{\mathrm{loop}}-\mathbf{x}^*_{\mathrm{loop}})
        -\bigl(
            S_\theta(\mathbf{x}_{\mathrm{loop}},\mathbf{u})
            -S_\theta(\mathbf{x}^*_{\mathrm{loop}},\mathbf{u})
         \bigr)
      \Bigr\|_2 \\[2pt]
    &\geq(1-q)\bigl\|
        \mathbf{x}_{\mathrm{loop}}-\mathbf{x}^*_{\mathrm{loop}}
      \bigr\|_2 ,
\end{align}
which is Eq.~\ref{eq:app-growth} after squaring. Consequently the sublevel set
$\{E\leq c\}$ confines $\mathbf{x}_{\mathrm{loop}}$ to a ball, on which the continuous
map $d$ is bounded, and the decoder term of Eq.~\ref{eq:app-split} then bounds
$\mathbf{x}_{\mathcal{O}}$ through
$\|\mathbf{x}_{\mathcal{O}}-d(\mathbf{x}_L)\|_2\leq\sqrt{2c}$; the sublevel
set is therefore bounded.

For the flow, $\tfrac{\mathrm{d}}{\mathrm{d}t}E=-\|\nabla_{\mathbf{x}}E\|_2^2\leq0$,
so a trajectory remains in the compact sublevel set of its initial value, and local
Lipschitz continuity of the gradient yields a solution for all $t\geq0$. Its
$\omega$-limit set is then nonempty, compact and invariant, and the energy is constant
on it, being the monotone limit of $E(\mathbf{x}(t))$; along the trajectory issued
from any of its points the energy is therefore constant, forcing
$\nabla_{\mathbf{x}}E=0$ there. By Theorem~\ref{thm:unique-app} the $\omega$-limit set
is $\{\mathbf{x}^*\}$, and the trajectory converges to it.
\end{proof}

\begin{proposition}[Nondegeneracy of the free equilibrium]
\label{prop:nondeg}
Let the maps of Eq.~\ref{eq:phi} and the decoder be twice continuously differentiable, and let $\mathbf{x}^*$ be a state of zero energy. No condition is placed on the Lipschitz constants of the maps.
\begin{enumerate}[label=(\roman*),leftmargin=*,topsep=2pt,itemsep=1pt]
    \item The Hessian $\mathbf{H}=\nabla^2_{\mathbf{x}\mathbf{x}}E(\theta,\mathbf{x}^*)$ in the free value nodes is positive definite if and only if $1\notin\operatorname{spec}\mathbf{J}_f(\mathbf{x}^*_L)$, and then $\lambda_{\min}(\mathbf{H})=\sigma_{\min}(\mathbf{J}_{\mathbf{R}})^2$, with $\mathbf{J}_{\mathbf{R}}$ as in Eq.~\ref{eq:app-JR}. The condition holds in particular if $\rho(\mathbf{J}_f(\mathbf{x}^*_L))<1$, and always under Assumption~\ref{ass:A1}, in which case $\mathbf{x}^*$ is the state of Theorem~\ref{thm:unique-app}.
    \item Under this condition, $\mathbf{x}^*$ is an isolated stationary point of $E(\theta,\cdot)$, hence an isolated global minimizer, and it is exponentially stable for the gradient flow.
    \item If $\rho\coloneqq\rho(\mathbf{J}_f(\mathbf{x}^*_L))<1$, then for every $\epsilon>0$ there are a neighborhood $U$ of $\mathbf{x}^*_L$ and a constant $c_\epsilon$ such that the iterates $\mathbf{z}^{k+1}=f_\theta(\mathbf{z}^k,\mathbf{u})$ with $\mathbf{z}^0\in U$ satisfy $\|\mathbf{z}^k-\mathbf{x}^*_L\|_2\leq c_\epsilon(\rho+\epsilon)^k\|\mathbf{z}^0-\mathbf{x}^*_L\|_2$.
\end{enumerate}
\end{proposition}

\begin{proof}
\emph{(i).} Write the full energy as $E=\tfrac12\|\mathbf{R}(\mathbf{x})\|_2^2$ with $\mathbf{R}(\mathbf{x})=\mathbf{x}-S^{\mathrm{full}}_\theta(\mathbf{x},\mathbf{u})$, where $S^{\mathrm{full}}_\theta(\cdot,\mathbf{u})$ appends the block $d(\mathbf{x}_L)$ to Eq.~\ref{eq:S}. Then
\begin{equation}
    \mathbf{H}=\mathbf{J}_{\mathbf{R}}^{\!\top}\mathbf{J}_{\mathbf{R}}+\sum_a R_a\,\nabla^2R_a ,
\end{equation}
and the second sum vanishes at $\mathbf{x}^*$, where $\mathbf{R}(\mathbf{x}^*)=0$. Hence $\mathbf{H}\succ0$ if and only if $\mathbf{J}_{\mathbf{R}}$ is invertible, and then $\lambda_{\min}(\mathbf{H})=\sigma_{\min}(\mathbf{J}_{\mathbf{R}})^2$. No vertex reads $\mathbf{x}_{\mathcal{O}}$, so in the block ordering $(\mathbf{x}_{\mathrm{loop}},\mathbf{x}_{\mathcal{O}})$,
\begin{equation}
    \mathbf{J}_{\mathbf{R}}=\mathbf{I}-\mathbf{J}_{S^{\mathrm{full}}}
    =\begin{pmatrix}
        \mathbf{I}-\mathbf{J}_S & \mathbf{0} \\[3pt]
        -\mathbf{D}             & \mathbf{I}
     \end{pmatrix},
     \qquad
     \mathbf{D}=\frac{\partial\,d(\mathbf{x}_L)}{\partial\mathbf{x}_{\mathrm{loop}}} ,
     \label{eq:app-JR}
\end{equation}
which is invertible if and only if $\mathbf{I}-\mathbf{J}_S(\mathbf{x}^*_{\mathrm{loop}})$ is. Solving $(\mathbf{I}-\mathbf{J}_S)\mathbf{v}=0$ blockwise gives $\mathbf{v}_1=\mathbf{J}_{\phi_1}(\mathbf{x}^*_L)\mathbf{v}_L$ and $\mathbf{v}_i=\mathbf{J}_{\phi_i}(\mathbf{x}^*_{i-1})\mathbf{v}_{i-1}$, so a nonzero solution exists if and only if $\mathbf{J}_{\phi_L}(\mathbf{x}^*_{L-1})\cdots\mathbf{J}_{\phi_1}(\mathbf{x}^*_L)$ has the eigenvalue $1$, with $\mathbf{v}_L$ as eigenvector. Since $\mathbf{x}^*_1=\phi_1(\mathbf{x}^*_L)$ and $\mathbf{x}^*_i=\phi_i(\mathbf{x}^*_{i-1})$ at a zero-energy state, this product is $\mathbf{J}_f(\mathbf{x}^*_L)$ by the chain rule applied to Eq.~\ref{eq:T}. If $\rho<1$, the eigenvalue $1$ is excluded. Under Assumption~\ref{ass:A1}, $\|\mathbf{J}_S\|_2\leq q<1$ by Lemma~\ref{lem:contraction}, so $\mathbf{I}-\mathbf{J}_S$ is invertible, and $\mathbf{x}^*$ is the unique zero-energy state of Theorem~\ref{thm:unique-app}.

\emph{(ii).} The map $\nabla_{\mathbf{x}}E(\theta,\cdot)$ is continuously differentiable with invertible Jacobian $\mathbf{H}$ at $\mathbf{x}^*$, so by the inverse function theorem it is injective on a neighborhood of $\mathbf{x}^*$, which therefore contains no other stationary point, and hence no other minimizer; $\mathbf{x}^*$ is a global minimizer because $E\geq0=E(\theta,\mathbf{x}^*)$. The linearization of the gradient flow at $\mathbf{x}^*$ is $\dot{\mathbf{v}}=-\mathbf{H}\mathbf{v}$, whose eigenvalues are negative, so $\mathbf{x}^*$ is exponentially stable by Lyapunov's indirect method.

\emph{(iii).} For $\epsilon>0$, there is a norm $\|\cdot\|_\epsilon$ on $\mathbb{R}^{d_L}$ whose induced operator norm satisfies $\|\mathbf{J}_f(\mathbf{x}^*_L)\|_\epsilon\leq\rho+\epsilon/2$. By continuity of $\mathbf{J}_f$, $\|\mathbf{J}_f\|_\epsilon\leq\rho+\epsilon<1$ on a closed $\|\cdot\|_\epsilon$-ball $U$ around $\mathbf{x}^*_L$, so by the mean value inequality $f_\theta(\cdot,\mathbf{u})$ maps $U$ into itself and contracts it with modulus $\rho+\epsilon$. The Euclidean bound follows by equivalence of norms; this is Ostrowski's theorem.

\end{proof}

Two remarks are in order, and the first concerns where the hypothesis has been placed.
Assumption~\ref{ass:A1} constrains each edge map of the loop separately, and not the
composition $f_\theta(\cdot,\mathbf{u})$, which is the object the deep equilibrium model
iterates and the one a reader would expect to see assumed contractive. The next remark
shows that this is not a matter of convenience.

\begin{remark}[Why the hypothesis is placed on the individual maps]
\label{rem:whyA1}
Contractivity of $f_\theta(\cdot,\mathbf{u})$ alone secures the unique fixed point, but not
the uniqueness of the stationary point of the energy, which is a statement about
arbitrary states rather than about consistent ones. Solving
$\bigl(\mathbf{I}-\mathbf{J}_S(\mathbf{x}_{\mathrm{loop}})\bigr)\mathbf{v}=0$ blockwise
gives $\mathbf{v}_1=\mathbf{J}_{\phi_1}(\mathbf{x}_L)\mathbf{v}_L$ and
$\mathbf{v}_i=\mathbf{J}_{\phi_i}(\mathbf{x}_{i-1})\mathbf{v}_{i-1}$, whence
\begin{equation}
    \mathbf{I}-\mathbf{J}_S(\mathbf{x}_{\mathrm{loop}})
    \text{ is singular}
    \iff
    \mathbf{I}
    -\mathbf{J}_{\phi_L}(\mathbf{x}_{L-1})\cdots\mathbf{J}_{\phi_1}(\mathbf{x}_L)
    \text{ is singular},
\end{equation}
a product of Jacobians evaluated at value nodes that bear no relation to one another
unless the state is already a fixed point. Bounding the Jacobian of the composition
along consistent states does not control that product, whereas
Assumption~\ref{ass:A1} bounds its norm by $q^L$ uniformly. The assumption is thus
strictly stronger than contractivity of $f_\theta(\cdot,\mathbf{u})$, which it implies with
modulus $q^L$.
\end{remark}

\begin{remark}[Scope]
\label{rem:scope-inference}
Assumption~\ref{ass:A1} guarantees convergence of the direct iteration
$\mathbf{x}_{\mathrm{loop}}^{k+1}
=S_\theta(\mathbf{x}_{\mathrm{loop}}^{k},\mathbf{u})$ at rate $q$, and of
$\mathbf{z}^{k+1}=f_\theta(\mathbf{z}^{k},\mathbf{u})$ at rate $q^L$, but not that
of an arbitrary accelerated fixed-point solver, nor that of a particular
discretization of the gradient flow, which requires in addition a step size adapted to
the curvature of the energy. The correspondence established above is one of terminal
states and not of trajectories: these procedures traverse the state space differently,
yet whenever one of them settles at a stationary solution of its own equilibrium
problem, that solution is $\mathbf{x}^*$.
\end{remark}

\subsection{Learning}
\label{app:learning}

We now turn to the parameter update. Let $\mathbf{x}^*(\theta)$ denote a free
equilibrium, $\nabla_{\mathbf{x}}E(\theta,\mathbf{x}^*)=0$, and let
\begin{equation}
    \mathcal{L}(\theta)
    =C\bigl(\mathbf{x}^*_{\mathcal{O}}(\theta),\mathbf{y}\bigr)
\end{equation}
be the supervised cost evaluated there, which is the quantity the implicit gradient
differentiates. We regard $C$ as a function of the whole free state, constant in every
block but the output one, so that $\nabla_{\mathbf{x}}C$ is supported on the output
vertex; this is the precise sense in which the nudge is a force applied at the output
alone. For $\beta$ in a neighborhood of the origin, the nudged equilibrium
$\mathbf{x}^\beta$ satisfies
$\nabla_{\mathbf{x}}\bigl[
    E(\theta,\mathbf{x}^\beta)+\beta\,C(\mathbf{x}^\beta,\mathbf{y})
 \bigr]=0$,
and all gradients in $\theta$ are understood as in Remark~\ref{rem:encoder}.

\begin{theorem}[Formal version of Theorem 2]
\label{thm:ep-app}
Let $\mathbf{x}^*$ be a free equilibrium at which $\mathbf{H}=\nabla^2_{\mathbf{x}\mathbf{x}}E(\theta,\mathbf{x}^*)\succ0$, let $E$ be twice continuously differentiable jointly in $(\theta,\mathbf{x})$ on a neighborhood of $(\theta,\mathbf{x}^*)$, and let $C$ be twice continuously differentiable in $\mathbf{x}$ on a neighborhood of $\mathbf{x}^*$. No condition is placed on the Lipschitz constants of the maps.
\begin{enumerate}[label=(\roman*),leftmargin=*,topsep=2pt,itemsep=1pt]
    \item \emph{Exact phases.} If both phases are solved exactly, $\beta\mapsto\mathbf{x}^\beta$ is well defined and continuously differentiable near $\beta=0$, and
    \begin{equation}
        \lim_{\beta\to0}G_{\mathrm{EP}}(\beta)
        =\lim_{\beta\to0}\frac{\nabla_\theta E(\theta,\mathbf{x}^\beta)-\nabla_\theta E(\theta,\mathbf{x}^*)}{\beta}
        =\nabla_\theta\mathcal{L}(\theta) .
        \label{eq:app-main-limit}
    \end{equation}
    If moreover $E$ is $C^3$ jointly and $C$ is $C^3$ in $\mathbf{x}$, then $G_{\mathrm{EP}}(\beta)=\nabla_\theta\mathcal{L}(\theta)+O(\beta)$; if both are $C^4$, the centered estimator $G^{\mathrm{c}}_{\mathrm{EP}}(\beta)\coloneqq(2\beta)^{-1}\bigl[\nabla_\theta E(\theta,\mathbf{x}^{+\beta})-\nabla_\theta E(\theta,\mathbf{x}^{-\beta})\bigr]$ satisfies $G^{\mathrm{c}}_{\mathrm{EP}}(\beta)=\nabla_\theta\mathcal{L}(\theta)+O(\beta^2)$.
    \item \emph{Zero-energy equilibria.} If $\mathbf{x}^*$ has zero energy and the maps of Eq.~\ref{eq:phi} and the decoder are twice continuously differentiable jointly in $\theta$ and in their arguments, the hypothesis $\mathbf{H}\succ0$ is equivalent to $1\notin\operatorname{spec}\mathbf{J}_f(\mathbf{x}^*_L)$ (Proposition~\ref{prop:nondeg}), and $\nabla_\theta\mathcal{L}(\theta)$ is the implicit gradient of the deep equilibrium model of Section~\ref{sec:deq} at its fixed point $\mathbf{x}^*_L$.
    \item \emph{Truncated phases.} Let $\lambda\coloneqq\lambda_{\min}(\mathbf{H})$. There are $\delta,\beta_0>0$ and $M<\infty$ such that, for $0<|\beta|<\beta_0$ and any states $\tilde{\mathbf{x}}^*,\tilde{\mathbf{x}}^\beta$ within distance $\delta$ of $\mathbf{x}^*$, the estimator $\tilde G_{\mathrm{EP}}(\beta)=\beta^{-1}\bigl[\nabla_\theta E(\theta,\tilde{\mathbf{x}}^\beta)-\nabla_\theta E(\theta,\tilde{\mathbf{x}}^*)\bigr]$ satisfies
    \begin{equation}
        \bigl\|\tilde G_{\mathrm{EP}}(\beta)-G_{\mathrm{EP}}(\beta)\bigr\|_2\leq\frac{4M}{\lambda|\beta|}\,(r_0+r_\beta),
        \quad
        r_0\coloneqq\|\nabla_{\mathbf{x}}E(\theta,\tilde{\mathbf{x}}^*)\|_2,
        \quad
        r_\beta\coloneqq\|\nabla_{\mathbf{x}}F_\beta(\theta,\tilde{\mathbf{x}}^\beta,\mathbf{y})\|_2 .
        \label{eq:app-truncated}
    \end{equation}
    The same holds for the centered estimator with $\tfrac12(r_\beta+r_{-\beta})$ in place of $r_0+r_\beta$ and, at a zero-energy $\mathbf{x}^*$, for $\beta^{-1}\nabla_\theta E(\theta,\tilde{\mathbf{x}}^\beta)$ with $r_\beta$ alone. If the nudged states are replaced by $\mathbf{x}^*\pm\beta\tilde{\mathbf{v}}$, where $\tilde{\mathbf{v}}$ solves $\mathbf{H}\mathbf{v}=-\nabla_{\mathbf{x}}C$ up to a residual $r_{\mathrm{lin}}$, the centered estimator is within $Mr_{\mathrm{lin}}/\lambda+O(\beta^2)$ of $\nabla_\theta\mathcal{L}(\theta)$, under the $C^4$ assumption of (i).
\end{enumerate}
\end{theorem}

\begin{proof}
\emph{Step 1: the nudged branch.}
Define
\begin{equation}
    \Phi(\mathbf{x},\beta)
    \coloneqq
    \nabla_{\mathbf{x}}E(\theta,\mathbf{x})
    +\beta\,\nabla_{\mathbf{x}}C(\mathbf{x},\mathbf{y}) ,
\end{equation}
continuously differentiable in $(\mathbf{x},\beta)$ by hypothesis and vanishing at
$(\mathbf{x}^*,0)$. Its partial Jacobian in $\mathbf{x}$ at that point is
$\mathbf{H}$, invertible by positive definiteness, so the implicit function theorem
provides a neighborhood of $\beta=0$ carrying a unique continuously differentiable
branch $\beta\mapsto\mathbf{x}^\beta$ with $\Phi(\mathbf{x}^\beta,\beta)=0$, equal to
$\mathbf{x}^*$ at the origin. If $E$ and $C$ are $C^{k+1}$, then $\Phi$ is $C^{k}$ and
so is the branch.

\emph{Step 2: the response to the nudge.}
Differentiating $\Phi(\mathbf{x}^\beta,\beta)=0$ with respect to $\beta$ at $\beta=0$,
\begin{equation}
    \mathbf{H}\,\dot{\mathbf{x}}^*
    +\nabla_{\mathbf{x}}C(\mathbf{x}^*,\mathbf{y})=0 ,
    \qquad\text{that is}\qquad
    \dot{\mathbf{x}}^*
    \coloneqq
    \frac{\mathrm{d}\mathbf{x}^\beta}{\mathrm{d}\beta}\bigg|_{\beta=0}
    =-\mathbf{H}^{-1}\nabla_{\mathbf{x}}C .
    \label{eq:app-response}
\end{equation}

\emph{Step 3: the implicit gradient.}
Since $\nabla_{\mathbf{x}}E$ is continuously differentiable jointly and $\mathbf{H}$
is invertible, the same theorem applied to
$\nabla_{\mathbf{x}}E(\theta,\mathbf{x}^*(\theta))=0$ in the variable $\theta$ yields
a differentiable equilibrium branch with
$\partial\mathbf{x}^*/\partial\theta
=-\mathbf{H}^{-1}\nabla^2_{\mathbf{x}\theta}E$, the matrix
$\nabla^2_{\mathbf{x}\theta}E$ having entries
$\partial^2E/\partial x_a\partial\theta_b$. The chain rule and the symmetry of
$\mathbf{H}$ then give
\begin{equation}
    \nabla_\theta\mathcal{L}
    =\Bigl(\frac{\partial\mathbf{x}^*}{\partial\theta}\Bigr)^{\!\top}
     \nabla_{\mathbf{x}}C
    =-\bigl(\nabla^2_{\mathbf{x}\theta}E\bigr)^{\!\top}
      \mathbf{H}^{-1}\nabla_{\mathbf{x}}C .
    \label{eq:app-implicit}
\end{equation}

\emph{Step 4: the estimator.}
The numerator of $G_{\mathrm{EP}}(\beta)$ is a difference quotient of
$u(\beta)\coloneqq\nabla_\theta E(\theta,\mathbf{x}^\beta)$, differentiable at the
origin by Step~1. By the chain rule and Eq.~\ref{eq:app-response},
\begin{equation}
    \lim_{\beta\to0}G_{\mathrm{EP}}(\beta)
    =u'(0)
    =\nabla^2_{\theta\mathbf{x}}E\;\dot{\mathbf{x}}^*
    =-\nabla^2_{\theta\mathbf{x}}E\,\mathbf{H}^{-1}\nabla_{\mathbf{x}}C .
    \label{eq:app-ep}
\end{equation}
As $E$ is twice continuously differentiable, the symmetry of second derivatives gives
$\nabla^2_{\theta\mathbf{x}}E=(\nabla^2_{\mathbf{x}\theta}E)^{\!\top}$, so
Eqs.~\ref{eq:app-ep} and \ref{eq:app-implicit} coincide, which proves
Eq.~\ref{eq:app-main-limit}.

\emph{Step 5: the orders.}
Under the stated additional smoothness, $u$ is $C^{2}$, respectively $C^{3}$, by
Step~1. Taylor expansion about the origin gives
\begin{align}
    u(\beta)-u(0)
    &=\beta\,u'(0)+\tfrac{1}{2}\beta^{2}u''(0)+o(\beta^{2}) , \\[2pt]
    u(\beta)-u(-\beta)
    &=2\beta\,u'(0)+\tfrac{1}{3}\beta^{3}u'''(0)+o(\beta^{3}) ,
\end{align}
the even-order terms cancelling by symmetry in the second line. Dividing by $\beta$
and by $2\beta$ respectively yields
$G_{\mathrm{EP}}(\beta)=u'(0)+O(\beta)$ and
$G^{\mathrm{c}}_{\mathrm{EP}}(\beta)=u'(0)+O(\beta^{2})$.

\emph{Step 6: zero-energy equilibria.} The equivalence is Proposition~\ref{prop:nondeg}(i). Under it, $\mathbf{I}-\mathbf{J}_f(\mathbf{x}^*_L)$ is invertible, and the implicit function theorem applied to $\mathbf{z}-f_{\theta'}(\mathbf{z},\mathbf{u})=0$ gives a continuously differentiable branch $\theta'\mapsto\mathbf{z}(\theta')$ of fixed points through $\mathbf{x}^*_L$; the gradient of $\theta'\mapsto C(d_{\theta'}(\mathbf{z}(\theta')),\mathbf{y})$ at $\theta$ is the implicit gradient of Section~\ref{sec:deq}. Lifted through Lemma~\ref{lem:fp}, with $\mathbf{x}_{\mathcal{O}}=d_{\theta'}(\mathbf{z}(\theta'))$, this branch consists of zero-energy, hence stationary, states through $\mathbf{x}^*$. The branch of Step~3 is the unique branch of stationary states through $\mathbf{x}^*$ near $(\theta,\mathbf{x}^*)$, so the two coincide, and $\mathcal{L}(\theta')=C(d_{\theta'}(\mathbf{z}(\theta')),\mathbf{y})$ for $\theta'$ near $\theta$.

\emph{Step 7: truncated phases.} Let $B$ be the closed ball of radius $\delta$ around $\mathbf{x}^*$, with $\delta$ small enough that $\nabla^2_{\mathbf{x}\mathbf{x}}E\succeq\tfrac{\lambda}{2}\mathbf{I}$ on $B$, and $\beta_0$ small enough that $\beta_0\sup_B\|\nabla^2_{\mathbf{x}\mathbf{x}}C\|_2\leq\tfrac{\lambda}{4}$ and that the branch $\mathbf{x}^{\pm\beta}$ of Step~1 lies in $B$ for $|\beta|<\beta_0$; both are possible by continuity. Then $E$ and $F_{\pm\beta}$ are $\tfrac{\lambda}{2}$- and $\tfrac{\lambda}{4}$-strongly convex on the convex set $B$, and a $\mu$-strongly convex function $f$ satisfies $\|\nabla f(\mathbf{a})-\nabla f(\mathbf{b})\|_2\geq\mu\|\mathbf{a}-\mathbf{b}\|_2$ there. Taking $\mathbf{b}$ at the exact equilibria, where the gradients vanish, gives $\|\tilde{\mathbf{x}}^*-\mathbf{x}^*\|_2\leq2r_0/\lambda$ and $\|\tilde{\mathbf{x}}^{\beta}-\mathbf{x}^{\beta}\|_2\leq4r_\beta/\lambda$. With $M\coloneqq\sup_B\|\nabla^2_{\theta\mathbf{x}}E\|_2$, the map $\mathbf{x}\mapsto\nabla_\theta E(\theta,\mathbf{x})$ is $M$-Lipschitz on $B$, so $\|\tilde G_{\mathrm{EP}}(\beta)-G_{\mathrm{EP}}(\beta)\|_2\leq M|\beta|^{-1}\bigl(\|\tilde{\mathbf{x}}^{\beta}-\mathbf{x}^{\beta}\|_2+\|\tilde{\mathbf{x}}^*-\mathbf{x}^*\|_2\bigr)$, which gives Eq.~\ref{eq:app-truncated}. The centered estimator is handled in the same way, its denominator contributing a factor $\tfrac12$. At a zero-energy $\mathbf{x}^*$, $\nabla_\theta E(\theta,\mathbf{x}^*)=0$, as every term of this gradient carries a factor $\boldsymbol{\varepsilon}_i$, so omitting the free term costs nothing. For the linear response, $\mathbf{H}\succeq\lambda\mathbf{I}$ gives $\|\tilde{\mathbf{v}}-\dot{\mathbf{x}}^*\|_2\leq r_{\mathrm{lin}}/\lambda$ by Eq.~\ref{eq:app-response}; the centered difference of $\beta\mapsto\nabla_\theta E(\theta,\mathbf{x}^*+\beta\tilde{\mathbf{v}})$ equals $\nabla^2_{\theta\mathbf{x}}E\,\tilde{\mathbf{v}}+O(\beta^2)$, while $\nabla^2_{\theta\mathbf{x}}E\,\dot{\mathbf{x}}^*=\nabla_\theta\mathcal{L}$ by Step~4.
\end{proof}

\end{document}